\documentclass{article}
\usepackage{graphicx} % Required for inserting images
\usepackage{times}
\usepackage{soul}
\usepackage{url}
\usepackage[hidelinks]{hyperref}
\usepackage[utf8]{inputenc}
\usepackage[small]{caption}
\usepackage{graphicx}
\usepackage{amsmath}
\usepackage{amsthm}
\usepackage{amssymb}
\usepackage{booktabs}
\usepackage{algorithm}
\usepackage{algpseudocode}
\usepackage{longtable}
\usepackage{array}
\usepackage{geometry}
\usepackage{multirow}
\usepackage{xcolor}
\usepackage{etoolbox}
\usepackage{colortbl}
\usepackage{tabularx}
\usepackage{placeins}
\usepackage{xcolor}
\usepackage{makecell}
\usepackage{tikz}
\usepackage{tablefootnote}
\usetikzlibrary{shapes.geometric, arrows.meta, positioning, calc, shadows}
\usepackage[numbers,sort&compress]{natbib}
\usepackage{pgfplots}
\pgfplotsset{compat=1.18}
\usepgfplotslibrary{groupplots}
\usepackage{xspace}
\newcommand{\AMA}{Ambiguity Mitigation Agent\xspace}
\newcommand{\AMAtab}{\makecell[l]{Ambiguity\\Mitigation Agent}}
\newcommand{\AMAtabstar}{\makecell[l]{Ambiguity\\Mitigation Agent $\star$}}

\definecolor{sectiongray}{gray}{0.9}
\usepackage{colortbl}

\usepackage[switch]{lineno}

\title{A knowledge-guided agentic framework for mitigating patient-context ambiguity in health queries}
\author{Mahyar Abbasian, Saba A. Farahani, Arshia Ilaty, Iman Azimi,\\
Hung Cao, Ramesh Jain, Amir M. Rahmani}
\usepackage{authblk}

\author[1,$\dagger$,*]{Mahyar Abbasian}
\author[1,$\dagger$]{Saba A. Farahani}
\author[1]{Arshia Ilaty}
\author[2]{Iman Azimi}
\author[1]{Hung Cao}
\author[1]{Ramesh Jain}
\author[1]{Amir M. Rahmani}
\affil[1]{University of California, Irvine, CA, USA}
\affil[2]{University of Turku, Turku, Finland}
\affil[$\dagger$]{These authors contributed equally to this work.}
\affil[*]{Corresponding author: abbasiam@uci.edu}
\date{January 2026}

\begin{document}

\maketitle

\begin{abstract}
Large language model (LLM)-based chatbots are increasingly used to seek
health information, yet their effectiveness depends on how they handle
ambiguity in user queries. Patient-context ambiguity, caused by undisclosed
patient-specific information, has received less attention and often leads
to generic or confusing responses. We propose the \AMA, a knowledge-guided
agent positioned between users and LLM chatbots that identifies and
resolves contextual ambiguity through targeted clarification questions. We
evaluated it across five downstream LLMs on two benchmarks, symptom-based
diagnosis retrieval (1,034 queries) and dietary-safety classification (487
queries), against two baseline methods. In diagnosis retrieval, exact Top-1
accuracy ranged from 57.7\% to 71.1\%, compared with at most 11.0\% for the
baselines. In dietary-safety classification, the agent improved accuracy
for all five models and achieved the highest Matthews correlation
coefficient for four. Addressing patient-context ambiguity can substantially
improve LLM-based health chatbots.
\end{abstract}

% #################### INTRODUCTION ####################

\section{Introduction}

There is a dramatically increasing use of large language models (LLMs) in healthcare, particularly in patient-facing applications such as chatbots ~\cite{Singhal2023MedPaLM,Thirunavukarasu2023LLMMedicine,Liu2024MedicalLLMSurvey,abbasian2025conversational}. Recent research has shown that these models have the potential to provide support for non-expert patients. This includes assisting users in seeking medical information, advice, and guidance to manage their health conditions \cite{abbasian2025conversational, Ayers2023Chatbot, Szymanski2024NutritionLLM}.

In practice, however, this promise depends on how users communicate and how the LLM interprets their queries. User-chatbot interactions often include brief, open-ended queries about health concerns, symptoms, or treatment options. The user's knowledge about the topic may be limited, incomplete, or inaccurate, which can contribute to the ambiguity of their queries. Therefore, the effectiveness and trustworthiness of the LLM responses heavily depend on the model's ability to recognize and address the ambiguity inherent in patient queries~\cite{Draelos2026HealthAdvice,GurArieh2026Ambiguity}.

There is a large body of literature that examines ambiguity in LLMs. Existing work has focused mainly on three types of ambiguity. First, \textit{linguistic ambiguity} is the most surface-level form, arising from the wording of the query itself \cite{Piantadosi2012Ambiguity,Liu2023,Min2020}. Several studies focus on linguistic ambiguity, targeting the resolution of polysemy, semantic underspecification, and lexical vagueness within user queries~\cite{Naik2025}. For example, Liu et al.~\cite{Liu2023} showed that even the strongest LLMs fail to recognize and characterize lexical, syntactic and scopal ambiguity, and Deng et al.~\cite{Deng2024} proposed having the model rewrite the query into a clearer form before answering. These approaches improve linguistic clarity, but rewriting or restructuring the user query cannot recover missing information.

Second, \textit{knowledge ambiguity} arises when the model's uncertainty is due to incomplete or missing knowledge ~\cite{Mallen2023ParametricNonParametric,Kadavath2022,Zhang2025MedicalQAUnifying}. It includes entity ambiguity, where a medical term or acronym maps to multiple possible referents, temporal knowledge staleness, and gaps in specialized domain knowledge \cite{Yin2023}. Retrieval-Augmented Generation (RAG) \cite{Lewis2020RAG} has been proposed to mitigate this ambiguity by augmenting the model with external knowledge sources during inference time. Structured medical knowledge graphs (KGs) have also been integrated with LLMs for diagnosis prediction~\cite{Gao2025DRKNOWS} and medical question answering~\cite{Rezaei2025AMGRAG}. KG-based approaches close knowledge gaps. However, a knowledge base has no access to the user medical history that they simply did not provide.

Third, \textit{intent ambiguity} is the uncertainty arising from the model's inability to infer the user's goal behind their query \cite{Tamkin2023,Zhang2024,Niwa2024}. Intent ambiguity has been explored in the context of task-oriented dialogue, where systems should infer the user's objective over multiple turns \cite{HosseiniAsl2020SimpleTOD,Heck2020TripPy}. The existing literature focuses on proactive clarification strategies. Zhang and Choi~\cite{Zhang2023} and Kuhn et al.~\cite{Kuhn2023} developed approaches that ask follow-up questions when the user's intent is unclear. In another work, Kim et al.~\cite{Kim2023} proposed Tree of Clarifications to represent the space of possible user goals explicitly. These methods help the model to infer what the user intends. However, resolving intent ambiguity is insufficient when the ambiguity stems from missing user-specific context.

Within healthcare chatbots, \textit{contextual ambiguity} has received relatively less attention compared to the other ambiguity types. The missing information is in unobserved or undisclosed patient-specific context. For example, a patient may ask, "What should I take for a headache?" without providing medical history, current medications, or lifestyle factors~\cite{Modderkolk2026Reality}. In such scenarios, existing state-of-the-art LLMs often fail to identify missing context or seek clarifying information~\cite{Li2024MediQ,Draelos2026HealthAdvice}, instead providing generic responses or a list of plausible answers with several caveats. Unfortunately, this behavior may overwhelm, mislead, or confuse patients, potentially leading to inappropriate self-management decisions or adverse health outcomes. The aim of this paper is to tackle contextual ambiguity in healthcare chatbots.

In this paper, we propose the \AMA, a knowledge-guided agentic approach that enables LLMs to handle contextual ambiguity in patient-facing healthcare applications. The \AMA is an intermediary layer positioned between users and LLM chatbots. Guided by a task-specific knowledge graph, it identifies the plausible answers a query could support and the undisclosed patient context that would distinguish among them, acquires that context through a small number of targeted questions, and passes the query together with the acquired context to the downstream model. Our contributions are threefold. First, we introduce a knowledge-guided agent that treats clarification as hypothesis elimination over a knowledge graph, separating the decision of what to ask from the language model that words the question and generates the response. Second, we curate two controlled ambiguity-mitigation benchmarks with different output structures, symptom-based diagnosis retrieval and dietary-safety classification, in which the withheld patient context is recoverable only through clarification. Third, we evaluate the agent across five downstream LLMs along three dimensions: the accuracy of the downstream response, the consistency of that accuracy across downstream models, and the uncertainty of the downstream output under repeated generation.

% #################### RESULTS ####################
\section{Results}

We evaluated the \AMA on two controlled benchmarks: symptom-based diagnosis retrieval (1,034 cases spanning 32
reference diagnoses) and dietary-safety classification (487 queries
spanning 38 patient health contexts). In both benchmarks, the intended
answer could not be determined from the information initially presented
to the model, and the withheld patient-specific information could be
recovered only through clarification (Methods;
Table~\ref{tab:task_overview}). The \AMA was evaluated using five
downstream language models (GPT-5.5, Claude Opus 4.8, Gemini 3.1 Pro,
LLaMA 3.3 70B and Mistral Large) and compared with two configurations
that acquired no additional patient information: direct answering of the
underspecified query (\textsc{Basic}) and Rephrase-and-Respond
(\textsc{RaR}). For \textsc{RaR} the user's query was rephrased before answering.
We report the performance of the \AMA along three dimensions: the
accuracy of the downstream response, the consistency of that performance
across downstream models, and the uncertainty of the downstream output
under repeated generation. Each dimension was assessed in both the
diagnosis and the dietary-safety domain.

\subsection{Response accuracy of the Ambiguity Mitigation Agent}
\label{sec:accuracy_results}

The \AMA improved the accuracy of downstream responses in both tasks
(Tables~\ref{tab:diagnosis_results} and~\ref{tab:food_results}). It
outperformed on every diagnosis-retrieval measure for all
five downstream models. Moreover, in dietary-safety classification, the proposed agent achieved
the highest accuracy for all five models and the highest Matthews
correlation coefficient (MCC) for four. In the diagnosis task, accuracy
denotes retrieval of the reference diagnosis at rank 1 (Top-1) or within
the five returned diagnoses (Recall@5), assessed at broad
diagnostic-category, ICD-grouping and exact-diagnosis levels. In the
dietary-safety task, accuracy denotes agreement between the safety
decision and the reference label, that is, whether the recommendation was
appropriate for the patient's undisclosed condition. Because the label
distribution was imbalanced, MCC is the primary summary measure, and
\textsc{Not OK} is the positive class so that recall and the
false-negative rate reflect the identification of foods carrying a
relevant health risk (Methods).

\paragraph{Diagnosis retrieval.}
\label{sec:diagnosis_results}

Ambiguity mitigation by the \AMA substantially improved diagnosis retrieval
across all five downstream language models, with the largest gains
observed at the ICD and exact-diagnosis levels
(Table~\ref{tab:diagnosis_results}). The baseline configurations often
identified the correct broad diagnostic category, but their performance
declined markedly when a more specific diagnosis was required.

For exact Top-1 retrieval, overall accuracy of the \AMA ranged from
57.7\% to 71.1\%, compared with 0.6\% to 11.0\% for the \textsc{Basic}
and \textsc{RaR} configurations. The same pattern was observed for exact
Recall@5. The \AMA included the reference diagnosis among its five
outputs in 90.6\% to 90.9\% of all cases, whereas exact Recall@5 remained
between 2.0\% and 16.2\% for the baseline configurations.

Ambiguity mitigation also produced large improvements at the ICD level. Overall
ICD Top-1 accuracy ranged from 78.8\% to 80.6\% for the \AMA, compared
with 27.2\% to 45.9\% for the baselines. Overall ICD Recall@5 ranged from
91.6\% to 91.9\% for the \AMA, whereas the corresponding baseline values
ranged from 56.3\% to 67.5\%. These results indicated that the additional
symptom information improved both exact diagnosis selection and placement
of the correct diagnosis among the leading candidates.

At the broad-category level, the difference was smaller. Overall Top-1
accuracy of the \AMA ranged from 89.0\% to 90.8\%, compared with 74.7\%
to 83.8\% for the baselines. However, category Recall@5 was already high
without clarification, ranging from 92.9\% to 96.8\% across the baseline
configurations. For GPT-5.5 and Gemini 3.1 Pro, some baseline
configurations achieved higher category Recall@5 than the \AMA. This
pattern suggested that the initially visible symptoms were often
sufficient to identify a broadly relevant diagnostic family, while the
withheld information was more important for distinguishing among
diagnoses within that family.

The exact-label distribution was concentrated, with viral sinusitis,
acute viral pharyngitis and acute bronchitis accounting for 67.8\% of
the evaluation set. Viral sinusitis alone occurred in 300 of the 1,034
cases, corresponding to an exact Top-1 accuracy of 29.01\% for a constant
predictor that returned this diagnosis for every query. Every
\textsc{Basic} and \textsc{RaR} configuration performed below this
descriptive reference, whereas every \AMA configuration performed above
it. This comparison was used as a sanity check rather than as a
clinically meaningful baseline.

The \AMA abstained on 35 cases, representing 3.4\% of the evaluation
set. Because the abstention decision occurred before downstream language
generation, the same cases were unanswered for all five models; no
abstentions occurred in the dietary-safety task. Selective exact Top-1
accuracy ranged from 59.8\% to 73.6\%, compared with 57.7\% to 71.1\%
when abstentions were counted as incorrect. The limited difference
between selective and overall results showed that the observed
improvement was not produced by excluding a large proportion of
difficult cases.

\begin{table}[!htbp]
    \centering
    \caption{\textbf{Symptom-based diagnosis retrieval on the curated
    evaluation set ($n=1{,}034$).}
    Performance was evaluated at three levels of specificity: broad
    diagnostic category, ICD-level grouping, and exact diagnosis name.
    Selective results include only cases for which the \AMA returned
    an answer. Overall results include all evaluation cases, with
    abstentions counted as incorrect. Because the \textsc{Basic} and
    \textsc{RaR} configurations did not abstain, their selective and
    overall values are identical. Bold values indicate the best
    configuration within each downstream model and metric column.
    Underlined values indicate the highest value in the corresponding
    column across all models and configurations. Category and ICD values
    rely on reference mappings that were not independently
    clinician-adjudicated, whereas exact-name matching does not require
    these mappings. All values are percentages.}
    \label{tab:diagnosis_results}
    \scriptsize

    \resizebox{\textwidth}{!}{
    \begin{tabular}{llc|ccc|ccc|ccc|ccc}
        \toprule
        & & &
        \multicolumn{3}{c|}{\textbf{Top-1, selective}} &
        \multicolumn{3}{c|}{\textbf{Recall@5, selective}} &
        \multicolumn{3}{c|}{\textbf{Top-1, overall}} &
        \multicolumn{3}{c}{\textbf{Recall@5, overall}} \\

        \textbf{Model} &
        \textbf{Configuration} &
        \textbf{Abstain} &
        \textbf{Category} &
        \textbf{ICD} &
        \textbf{Exact} &
        \textbf{Category} &
        \textbf{ICD} &
        \textbf{Exact} &
        \textbf{Category} &
        \textbf{ICD} &
        \textbf{Exact} &
        \textbf{Category} &
        \textbf{ICD} &
        \textbf{Exact} \\
        \midrule

        GPT-5.5 &
        \AMAtab &
        3.4 &
        \textbf{92.1} &
        \textbf{82.2} &
        \textbf{64.3} &
        \textbf{96.9} &
        \textbf{95.1} &
        \textbf{\underline{94.1}} &
        \textbf{89.0} &
        \textbf{79.4} &
        \textbf{62.1} &
        \textbf{93.6} &
        \textbf{91.9} &
        \textbf{\underline{90.9}} \\

        GPT-5.5 &
        RaR &
        0.0 &
        80.3 &
        27.5 &
        1.7 &
        95.6 &
        60.5 &
        2.2 &
        80.3 &
        27.5 &
        1.7 &
        95.6 &
        60.5 &
        2.2 \\

        GPT-5.5 &
        Basic &
        0.0 &
        80.9 &
        27.2 &
        1.6 &
        95.6 &
        60.1 &
        2.0 &
        80.9 &
        27.2 &
        1.6 &
        95.6 &
        60.1 &
        2.0 \\

        \midrule

        Claude Opus 4.8 &
        \AMAtab &
        3.4 &
        \textbf{\underline{94.0}} &
        \textbf{81.6} &
        \textbf{72.6} &
        \textbf{96.7} &
        \textbf{94.9} &
        \textbf{93.9} &
        \textbf{\underline{90.8}} &
        \textbf{78.8} &
        \textbf{70.1} &
        \textbf{93.4} &
        \textbf{91.7} &
        \textbf{90.7} \\

        Claude Opus 4.8 &
        RaR &
        0.0 &
        79.2 &
        33.3 &
        4.5 &
        94.8 &
        63.4 &
        10.9 &
        79.2 &
        33.3 &
        4.5 &
        94.8 &
        63.4 &
        10.9 \\

        Claude Opus 4.8 &
        Basic &
        0.0 &
        75.2 &
        36.3 &
        1.5 &
        92.9 &
        67.2 &
        4.0 &
        75.2 &
        36.3 &
        1.5 &
        92.9 &
        67.2 &
        4.0 \\

        \midrule

        Gemini 3.1 Pro &
        \AMAtab &
        3.4 &
        \textbf{92.1} &
        \textbf{82.3} &
        \textbf{59.8} &
        \textbf{96.9} &
        \textbf{95.1} &
        \textbf{\underline{94.1}} &
        \textbf{89.0} &
        \textbf{79.5} &
        \textbf{57.7} &
        \textbf{93.6} &
        \textbf{91.9} &
        \textbf{\underline{90.9}} \\

        Gemini 3.1 Pro &
        RaR &
        0.0 &
        81.2 &
        29.3 &
        0.8 &
        96.6 &
        56.3 &
        3.3 &
        81.2 &
        29.3 &
        0.8 &
        96.6 &
        56.3 &
        3.3 \\

        Gemini 3.1 Pro &
        Basic &
        0.0 &
        83.8 &
        28.6 &
        0.6 &
        96.8 &
        67.5 &
        5.7 &
        83.8 &
        28.6 &
        0.6 &
        96.8 &
        67.5 &
        5.7 \\

        \midrule

        LLaMA 3.3 70B &
        \AMAtab &
        3.4 &
        \textbf{92.2} &
        \textbf{\underline{83.4}} &
        \textbf{\underline{73.6}} &
        \textbf{97.2} &
        \textbf{94.9} &
        \textbf{93.9} &
        \textbf{89.1} &
        \textbf{\underline{80.6}} &
        \textbf{\underline{71.1}} &
        \textbf{93.9} &
        \textbf{91.7} &
        \textbf{90.7} \\

        LLaMA 3.3 70B &
        RaR &
        0.0 &
        81.7 &
        44.5 &
        10.0 &
        94.9 &
        62.7 &
        15.7 &
        81.7 &
        44.5 &
        10.0 &
        94.9 &
        62.7 &
        15.7 \\

        LLaMA 3.3 70B &
        Basic &
        0.0 &
        81.1 &
        45.9 &
        11.0 &
        93.9 &
        62.1 &
        16.2 &
        81.1 &
        45.9 &
        11.0 &
        93.9 &
        62.1 &
        16.2 \\

        \midrule

        Mistral Large &
        \AMAtab &
        3.4 &
        \textbf{93.0} &
        \textbf{82.1} &
        \textbf{71.2} &
        \textbf{\underline{97.5}} &
        \textbf{94.8} &
        \textbf{93.8} &
        \textbf{89.8} &
        \textbf{79.3} &
        \textbf{68.8} &
        \textbf{\underline{94.2}} &
        \textbf{91.6} &
        \textbf{90.6} \\

        Mistral Large &
        RaR &
        0.0 &
        76.3 &
        37.1 &
        7.4 &
        94.7 &
        66.4 &
        12.3 &
        76.3 &
        37.1 &
        7.4 &
        94.7 &
        66.4 &
        12.3 \\

        Mistral Large &
        Basic &
        0.0 &
        74.7 &
        38.9 &
        2.7 &
        93.7 &
        66.3 &
        5.1 &
        74.7 &
        38.9 &
        2.7 &
        93.7 &
        66.3 &
        5.1 \\

        \bottomrule
    \end{tabular}}
\end{table}

\FloatBarrier
\paragraph{Dietary-safety classification.}
\label{sec:food_results}

Clarification by the \AMA generally improved dietary-safety
classification across the five downstream language models
(Table~\ref{tab:food_results}). The \AMA achieved the highest MCC for
GPT-5.5, Claude Opus 4.8, LLaMA 3.3 70B and Mistral Large, and the
highest accuracy for all five models. For Gemini 3.1 Pro, the
\textsc{Basic} configuration had the highest MCC, although the \AMA
achieved the highest accuracy and the lowest false-positive rate.

The effect of clarification was most apparent in the balance between
identifying unsafe foods and avoiding unnecessary restrictions. For
GPT-5.5 and Claude Opus 4.8, the \AMA substantially reduced
false-negative errors relative to both baseline configurations while
maintaining comparatively strong precision. For LLaMA 3.3 70B, the \AMA
and \textsc{Basic} had the same recall of 95.7\% and the same
false-negative count of eight, but the \AMA reduced the false-positive
rate from 23.6\% to 18.6\%. This increased precision from 71.5\% to
76.1\% and raised MCC from 0.701 to 0.750.

A similar pattern was observed for Mistral Large. The \AMA and
\textsc{Basic} had the same recall of 95.2\%, but the \AMA produced fewer
false positives and higher precision, resulting in a higher MCC. The
\textsc{RaR} configuration achieved the highest recall for this model,
but this gain was accompanied by a false-positive rate of 24.6\%, showing
that greater sensitivity alone did not provide the best overall
classification balance.

Gemini 3.1 Pro showed a different trade-off. The \textsc{Basic}
configuration achieved the highest recall and MCC, whereas the \AMA
produced the highest accuracy, the highest precision and the lowest
false-positive rate. This result indicated that clarification did not
dominate every metric for every model, but generally improved the balance
between detecting unsafe foods and avoiding unnecessary restrictions.

Rephrase-and-Respond did not provide a consistent advantage over direct
prompting. Its effect varied across models and often involved a trade-off
between recall and false-positive error. Overall, the \AMA produced the
best MCC for four of the five downstream models, while the Gemini results
showed that the benefit of clarification could depend on the model and
the relative importance assigned to false negatives and false positives.

\begin{table}[!htbp]
    \centering
    \caption{\textbf{Dietary safety classification on 487 queries.}
    The positive class is \textsc{Not OK}. A false negative denotes an
    unsafe food classified as safe, whereas a false positive denotes an
    unnecessary restriction. MCC is reported because the dataset contains
    301 \textsc{OK} and 186 \textsc{Not OK} queries. Bold values indicate
    the best result within each downstream-model block and metric column;
    $\star$ marks the best configuration for each model, ranked by MCC.
    Higher values are better for accuracy, MCC, F1, recall, and precision,
    whereas lower values are better for false-negative rate,
    false-positive rate, and false-negative count. Accuracy, recall,
    precision, and error rates are percentages.}
    \label{tab:food_results}
    \scriptsize
    \resizebox{\textwidth}{!}{
    \begin{tabular}{llrrrrrrrr}
        \toprule
        \textbf{Model} &
        \textbf{Configuration} &
        \textbf{Accuracy} &
        \textbf{MCC} &
        \textbf{F1} &
        \textbf{Recall} &
        \textbf{Precision} &
        \textbf{FN rate} &
        \textbf{FP rate} &
        \textbf{FN ($n$)} \\
        \midrule

        GPT-5.5 &
        \AMAtabstar &
        \textbf{89.3} &
        \textbf{0.783} &
        \textbf{0.869} &
        \textbf{92.5} &
        81.9 &
        \textbf{7.5} &
        12.6 &
        \textbf{14} \\

        GPT-5.5 &
        RaR &
        83.2 &
        0.640 &
        0.747 &
        65.1 &
        \textbf{87.7} &
        34.9 &
        \textbf{5.6} &
        65 \\

        GPT-5.5 &
        Basic &
        88.3 &
        0.757 &
        0.852 &
        88.2 &
        82.4 &
        11.8 &
        11.6 &
        22 \\

        \midrule

        Claude Opus 4.8 &
        \AMAtabstar &
        \textbf{91.8} &
        \textbf{0.837} &
        \textbf{0.900} &
        \textbf{96.8} &
        84.1 &
        \textbf{3.2} &
        11.3 &
        \textbf{6} \\

        Claude Opus 4.8 &
        RaR &
        83.4 &
        0.650 &
        0.741 &
        62.4 &
        \textbf{91.3} &
        37.6 &
        \textbf{3.7} &
        70 \\

        Claude Opus 4.8 &
        Basic &
        81.3 &
        0.607 &
        0.760 &
        77.4 &
        74.6 &
        22.6 &
        16.3 &
        42 \\

        \midrule

        Gemini 3.1 Pro &
        \AMAtab &
        \textbf{90.3} &
        0.795 &
        0.872 &
        86.0 &
        \textbf{88.4} &
        14.0 &
        \textbf{7.0} &
        26 \\

        Gemini 3.1 Pro &
        RaR &
        88.3 &
        0.751 &
        0.846 &
        83.9 &
        85.2 &
        16.1 &
        9.0 &
        30 \\

        Gemini 3.1 Pro &
        Basic $\star$ &
        89.5 &
        \textbf{0.799} &
        \textbf{0.877} &
        \textbf{97.3} &
        79.7 &
        \textbf{2.7} &
        15.3 &
        \textbf{5} \\

        \midrule

        LLaMA 3.3 70B &
        \AMAtabstar &
        \textbf{86.9} &
        \textbf{0.750} &
        \textbf{0.848} &
        \textbf{95.7} &
        \textbf{76.1} &
        \textbf{4.3} &
        \textbf{18.6} &
        \textbf{8} \\

        LLaMA 3.3 70B &
        RaR &
        85.6 &
        0.726 &
        0.834 &
        94.6 &
        74.6 &
        5.4 &
        19.9 &
        10 \\

        LLaMA 3.3 70B &
        Basic &
        83.8 &
        0.701 &
        0.818 &
        \textbf{95.7} &
        71.5 &
        \textbf{4.3} &
        23.6 &
        \textbf{8} \\

        \midrule

        Mistral Large &
        \AMAtabstar &
        \textbf{86.4} &
        \textbf{0.741} &
        \textbf{0.843} &
        95.2 &
        \textbf{75.6} &
        4.8 &
        \textbf{18.9} &
        9 \\

        Mistral Large &
        RaR &
        84.2 &
        0.718 &
        0.826 &
        \textbf{98.4} &
        71.2 &
        \textbf{1.6} &
        24.6 &
        \textbf{3} \\

        Mistral Large &
        Basic &
        85.0 &
        0.718 &
        0.829 &
        95.2 &
        73.4 &
        4.8 &
        21.3 &
        9 \\

        \bottomrule
    \end{tabular}}
\end{table}

\FloatBarrier
\subsection{Consistency of the Ambiguity Mitigation Agent across downstream models}
\label{sec:model_consistency}

\begin{figure*}[!t]
    \centering
    \includegraphics[width=\textwidth]{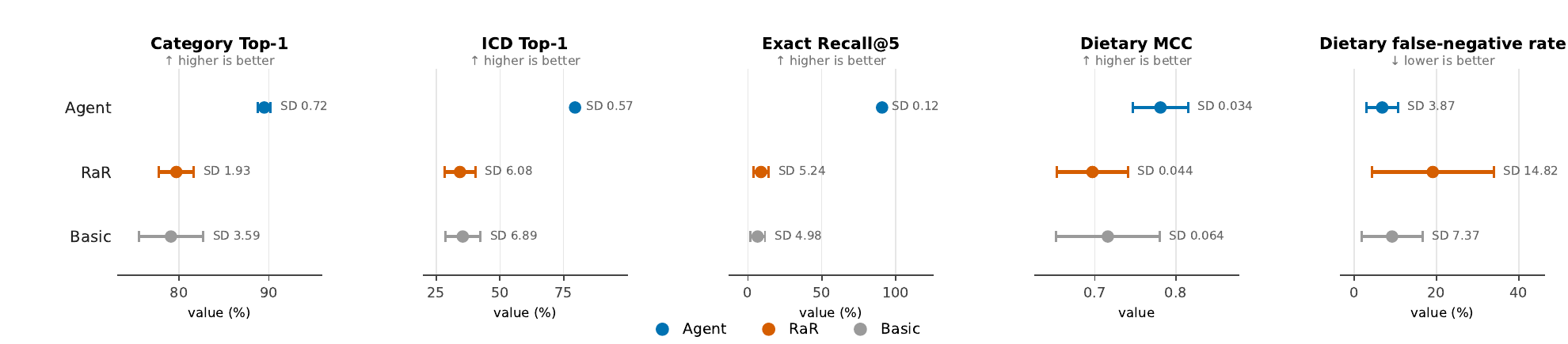}
    \caption{\textbf{Between-model performance and variation under each
    configuration.}
    Markers denote the mean across the five evaluated downstream language
    models, and horizontal error bars denote $\pm 1$ population standard
    deviation across the five model-specific values. Diagnosis metrics
    were computed over all 1,034 cases, with abstentions counted as
    incorrect, and dietary metrics were computed over all 487 queries.
    Higher values indicate better performance for category Top-1 accuracy,
    ICD Top-1 accuracy, exact Recall@5, and dietary MCC, whereas a lower
    value indicates better performance for the dietary false-negative
    rate, as shown by the arrows above each panel. For every displayed
    metric, the \AMA attained both a better mean and a smaller
    between-model standard deviation than either baseline, indicating
    stronger average performance and less dependence on the selected
    downstream model. MCC is unitless; all other quantities are expressed
    in percentage points. Exact Top-1 diagnosis accuracy was omitted
    because its near-zero baseline means made dispersion alone
    misleading.}
    \label{fig:model_dispersion}
\end{figure*}

We next examined whether the performance of each configuration depended
on which downstream language model was used. For each configuration and
metric, the five model-specific scores were summarized by their mean and
population standard deviation (SD) across the five models, so that a
smaller SD indicates that the five models produced more similar results
under that configuration (Methods). Because five models could show little
variation by all performing similarly poorly, SD was interpreted together
with the mean. In Figure~\ref{fig:model_dispersion}, each marker denotes
the mean across the five downstream models, each horizontal error bar
extends one SD on either side of the mean, and the arrows indicate
whether higher or lower values represent better performance for each
metric.

For overall category Top-1 diagnosis accuracy, the \AMA achieved a mean
of 89.5\% across the five models, with an SD of 0.72 percentage points.
The corresponding means were 79.7\% for \textsc{RaR} and 79.1\% for
\textsc{Basic}, with SDs of 1.93 and 3.59 percentage points,
respectively. Category-level accuracy was therefore both higher on
average and less dependent on the selected downstream model after
clarification.

The difference was more pronounced at the ICD level. Overall ICD Top-1
accuracy under the \AMA had a mean of 79.5\% and an SD of 0.57
percentage points. The corresponding means were 34.3\% for \textsc{RaR}
and 35.4\% for \textsc{Basic}, with SDs of 6.08 and 6.89 percentage
points. Thus, ICD-level performance varied substantially across
downstream models when the incomplete query was answered directly or
rephrased, whereas performance after clarification remained similar
across the model panel.

Exact Recall@5 showed the clearest diagnosis pattern. The \AMA achieved
a mean of 90.8\%, with an SD of only 0.12 percentage points. In
comparison, \textsc{RaR} achieved a mean of 8.9\% with an SD of 5.24
percentage points, while \textsc{Basic} achieved a mean of 6.6\% with an
SD of 4.98 percentage points. Clarification therefore combined
substantially higher exact retrieval performance with markedly smaller
differences among the five downstream models.

A similar pattern was observed for dietary safety. The \AMA achieved the
highest mean MCC, at 0.781, and the lowest SD, at 0.034. The
corresponding means were 0.697 for \textsc{RaR} and 0.716 for
\textsc{Basic}, with SDs of 0.044 and 0.064, respectively. The balance
between identifying unsafe foods and avoiding unnecessary restrictions
therefore varied less across models after clarification.

The dietary false-negative rate showed the same pattern in the favorable
lower direction. The \AMA had the lowest mean false-negative rate, at
6.8\%, and the lowest SD, at 3.87 percentage points. The corresponding
means were 19.1\% for \textsc{RaR} and 9.2\% for \textsc{Basic}, with
SDs of 14.82 and 7.37 percentage points. Both the average frequency of
missed unsafe foods and its dependence on the downstream model were
therefore lower under the \AMA.

Because the \AMA abstained on 35 diagnosis cases, the diagnosis analysis
was repeated using the same 999 answered cases for all three
configurations. The \AMA retained the lowest variation for the
representative category, ICD and Recall@5 measures. Its greater
cross-model consistency was therefore not explained by evaluating a
different or smaller set of cases.

Exact Top-1 diagnosis accuracy was the main exception and was omitted
from Figure~\ref{fig:model_dispersion}. Its between-model SD was higher
under the \AMA than under the baseline configurations. However, mean
exact Top-1 accuracy was 66.0\% for the \AMA, compared with only 3.5\%
for \textsc{Basic} and 4.9\% for \textsc{RaR}. The lower baseline
variation therefore occurred because all five baseline models performed
near zero, rather than because the baselines provided better or more
reliable exact diagnosis retrieval.

Overall, the \AMA produced stronger mean performance and lower
between-model variation for every metric shown in
Figure~\ref{fig:model_dispersion}, in both the diagnosis and the
dietary-safety domain. Performance under the \AMA was therefore less
dependent on the selected downstream language model than performance
under either baseline. This analysis was descriptive and applied only to
the five models included in the evaluation.

\FloatBarrier
\subsection{Uncertainty of the Ambiguity Mitigation Agent under repeated generation}
\label{sec:uncertainty_results}

We characterized how much the downstream output varied when the
same query was repeated. Two queries were selected from each task, and
each model--configuration pair was evaluated over 500 independent
repetitions per query. Output dispersion was quantified by predictive
entropy over the returned answers, where 0 bits indicates that the same
answer was returned on every repetition (Methods). For the \AMA, the
clarification trajectory, terminal state and final decision were
generated once and held fixed across repetitions, so the repeated runs
resampled only the downstream generation of an already clarified
decision; for \textsc{Basic} and \textsc{RaR}, the complete response was
regenerated on every repetition. The values reported below should
therefore be interpreted as observed output dispersion under these
sampling protocols rather than as a fully symmetric measure of
end-to-end uncertainty. Entropy characterizes consistency rather than
correctness: a low value means that a model repeatedly produced the same
answer, not that the answer agreed with the reference.

\paragraph{Diagnosis outputs.}
\label{sec:diagnosis_uncertainty}

Three quantities were reported for each model--configuration pair
(Table~\ref{tab:diagnosis_uncertainty}): $K$, the number of distinct
exact diagnosis strings returned across the 500 repetitions;
$H_{\mathrm{exact}}$, the predictive entropy over those exact strings;
and $H_{\mathrm{cat}}$, the entropy after the diagnosis strings were
mapped to clinical categories. The last measure separated variation in
wording from variation in the underlying clinical conclusion: a model
could have nonzero exact-string entropy while retaining zero
category-level entropy if it used different expressions for the same
clinical category.

Within this protocol, the \AMA produced substantially less
category-level variation than either baseline. Across the ten \AMA
item--model pairs, mean exact-string entropy was 0.170 bits and mean
category-level entropy was 0.017 bits. The corresponding mean
category-level entropies were 0.320 bits for \textsc{Basic} and 0.338
bits for \textsc{RaR}. The remaining variation under the \AMA therefore
occurred primarily in the exact wording of the diagnosis rather than in
the resolved clinical category.

This distinction was particularly clear for the ischemic heart disease
query. Under the \AMA, Mistral Large returned three distinct diagnosis
strings, resulting in an exact-string entropy of 1.145 bits, while all
outputs mapped to the same clinical category and produced a
category-level entropy of 0 bits. Claude Opus 4.8 showed a similar
pattern, with an exact-string entropy of 0.335 bits and a category-level
entropy of 0 bits. Variation in surface wording therefore did not
necessarily indicate variation in the clinical conclusion.

The baseline configurations showed greater dispersion on the same query.
They returned as many as 17 distinct diagnosis strings, and
category-level entropy reached 1.128 bits. For example, Gemini 3.1 Pro
under \textsc{RaR} returned 17 distinct diagnoses, with an exact-string
entropy of 2.703 bits and a category-level entropy of 0.930 bits,
indicating variation not only in wording but also in the clinical
category assigned across repetitions.

The two probe queries also showed that low entropy alone did not imply
correctness. For the acute viral pharyngitis query, several baseline
configurations returned acute tonsillitis on all 500 repetitions and
therefore had zero entropy but zero accuracy, because the same incorrect
diagnosis was returned consistently. The baseline configurations thus
exhibited both consistent error and variable error, whereas the \AMA
produced the correct modal diagnosis in every evaluated item--model pair
and showed minimal variation at the clinical-category level. Accuracy
was evaluated separately in the main analysis and was not used in
calculating $K$, $H_{\mathrm{exact}}$ or $H_{\mathrm{cat}}$.

\begin{table}[!htbp]
    \centering
    \caption{\textbf{Diagnosis-output uncertainty under repeated sampling.}
Two randomly selected diagnosis queries were evaluated using 500
independent repetitions for each model and configuration. $K$ denotes the
number of distinct exact diagnosis strings returned,
$H_{\mathrm{exact}}$ denotes predictive entropy over exact diagnosis
strings, and $H_{\mathrm{cat}}$ denotes entropy after mapping outputs to
their clinical categories. Item 0 had the reference diagnosis acute viral
pharyngitis, and Item 98 had the reference diagnosis ischemic heart
disease. In the \AMA condition, the symbolic clarification trajectory
and terminal hypothesis were held fixed, so the repetitions resampled
only downstream language generation. Outputs that could not be mapped to
a clinical category were retained as distinct categories rather than
excluded. Bold values identify the lowest-uncertainty configuration
within each model block, and $\star$ marks that configuration. Lower
values indicate less variation across repeated outputs. Entropy measures
the dispersion of the generated answers and does not, by itself, indicate
whether the resulting diagnosis was correct. Entropy is reported in
bits.}
    \label{tab:diagnosis_uncertainty}
    \small

    \resizebox{\textwidth}{!}{%
    \begin{tabular}{llrrr|rrr}
        \toprule
        & &
        \multicolumn{3}{c|}{\textbf{Item 0: acute viral pharyngitis}} &
        \multicolumn{3}{c}{\textbf{Item 98: ischemic heart disease}} \\
        \textbf{Model} &
        \textbf{Configuration} &
        \textbf{$K$} &
        \textbf{$H_{\mathrm{exact}}$} &
        \textbf{$H_{\mathrm{cat}}$} &
        \textbf{$K$} &
        \textbf{$H_{\mathrm{exact}}$} &
        \textbf{$H_{\mathrm{cat}}$} \\
        \midrule

        GPT-5.5 &
        \AMAtabstar &
        \textbf{1} & \textbf{0.000} & \textbf{0.000} &
        \textbf{1} & \textbf{0.000} & \textbf{0.000} \\
        GPT-5.5 &
        RaR &
        1 & 0.000 & 0.000 &
        6 & 1.006 & 0.988 \\
        GPT-5.5 &
        Basic &
        1 & 0.000 & 0.000 &
        5 & 1.232 & 1.128 \\
        \midrule

        Claude Opus 4.8 &
        \AMAtabstar &
        \textbf{1} & \textbf{0.000} & \textbf{0.000} &
        \textbf{2} & \textbf{0.335} & \textbf{0.000} \\
        Claude Opus 4.8 &
        RaR &
        11 & 2.092 & 0.000 &
        9 & 2.044 & 0.511 \\
        Claude Opus 4.8 &
        Basic &
        1 & 0.000 & 0.000 &
        16 & 1.161 & 0.932 \\
        \midrule

        Gemini 3.1 Pro &
        \AMAtabstar &
        \textbf{1} & \textbf{0.000} & \textbf{0.000} &
        \textbf{2} & \textbf{0.094} & \textbf{0.094} \\
        Gemini 3.1 Pro &
        RaR &
        3 & 0.058 & 0.000 &
        17 & 2.703 & 0.930 \\
        Gemini 3.1 Pro &
        Basic &
        2 & 0.053 & 0.000 &
        15 & 1.968 & 1.056 \\
        \midrule

        LLaMA 3.3 70B &
        \AMAtabstar &
        \textbf{3} & \textbf{0.074} & \textbf{0.021} &
        \textbf{2} & \textbf{0.053} & \textbf{0.053} \\
        LLaMA 3.3 70B &
        RaR &
        6 & 1.897 & 0.000 &
        7 & 1.208 & 0.205 \\
        LLaMA 3.3 70B &
        Basic &
        5 & 1.530 & 0.000 &
        5 & 0.847 & 0.021 \\
        \midrule

        Mistral Large &
        \AMAtabstar &
        \textbf{1} & \textbf{0.000} & \textbf{0.000} &
        \textbf{3} & 1.145 & \textbf{0.000} \\
        Mistral Large &
        RaR &
        3 & 0.236 & 0.000 &
        4 & 1.411 & 0.741 \\
        Mistral Large &
        Basic &
        1 & 0.000 & 0.000 &
        \textbf{3} & \textbf{0.344} & 0.067 \\
        \bottomrule
    \end{tabular}%
    }
\end{table}

\FloatBarrier
\paragraph{Dietary-safety decisions.}
\label{sec:food_uncertainty}

For the dietary-safety task, one \textsc{Not OK}-labeled and one
\textsc{OK}-labeled query were evaluated
(Table~\ref{tab:food_uncertainty}). Predictive entropy was calculated
separately for the two queries ($H_{\mathrm{unsafe}}$ and
$H_{\mathrm{safe}}$) and averaged to give the mean entropy; a value
approaching 1 bit indicates an almost even division between
\textsc{OK} and \textsc{Not OK} across the 500 repetitions. The 500
decisions from each query were pooled into a balanced set of 1,000
decisions per model--configuration pair to calculate the accompanying
classification measures, which describe whether the repeated decisions
were correct, whereas the entropy values describe how much the decisions
varied within each query.

Under this protocol, all five \AMA configurations returned the same
decision on every repetition for both queries, resulting in
$H_{\mathrm{unsafe}}=0$, $H_{\mathrm{safe}}=0$ and a mean entropy of
0 bits. The \AMA also classified both queries correctly in every
repetition. The zero-entropy result therefore reflected stable downstream
expression of the fixed clarified decision.

Variation in the baseline configurations occurred mainly on the
\textsc{Not OK}-labeled query, where entropy reached 0.343 bits for GPT-5.5
under \textsc{RaR}; 12 of the 15 model--configuration rows had zero entropy on
this query. On the \textsc{OK}-labeled query, 14 of the 15 rows had zero
entropy. The exception was Claude Opus 4.8 under \textsc{Basic}, with an
entropy of 0.021 bits (499 \textsc{OK} and one \textsc{Not OK}).

Rephrase-and-Respond did not reduce dispersion. Mean entropy was higher under \textsc{RaR} than under
\textsc{Basic} for GPT-5.5 (0.172 versus 0.000 bits), Claude Opus 4.8 (0.059
versus 0.010 bits) and Gemini 3.1 Pro (0.019 versus 0.000 bits), and was zero
under both configurations for LLaMA 3.3 70B and Mistral Large.

Taken together, dietary-safety decisions were highly consistent in all three
configurations. The \AMA returned the correct decision on every repetition for
all five models; \textsc{Basic} did so for GPT-5.5, Gemini 3.1 Pro, LLaMA 3.3
70B and Mistral Large, and \textsc{RaR} for LLaMA 3.3 70B and Mistral Large.
Because only two queries were examined, the \AMA decision was fixed before
repeated generation and one baseline row was conditional on resampled decisions,
these findings were treated as an exploratory analysis of downstream output
stability rather than an estimate of uncertainty across the full evaluation sets.

\begin{table}[!htbp]
    \centering
    \caption{\textbf{Dietary-safety uncertainty under repeated
    sampling.}
    One \textsc{Not OK}-labeled query and one \textsc{OK}-labeled query
    were each sampled 500 independent times per model and configuration,
    giving 1,000 binary decisions per row. The positive
    class was \textsc{Not OK}. $H_{\mathrm{unsafe}}$ and
    $H_{\mathrm{safe}}$ denote predictive entropy for the
    \textsc{Not OK}- and \textsc{OK}-labeled queries, respectively.
    Zero bits indicates a fully consistent output, while one bit indicates
    an approximately even split between the two labels. In the \AMA
    condition, the confirmed patient context and final safety decision
    were held fixed across repetitions, so the repetitions resampled only
    downstream language generation. Bold values indicate the best result
    within each model block and metric column. The $\star$ symbol marks
    the best-performing configuration for each model, ranked by MCC,
    then accuracy, then mean entropy; all tied configurations are marked.
    Low entropy indicates consistency, not
    necessarily correctness. Accuracy, recall, precision, and
    false-positive rate are percentages; MCC is unitless.}
    \label{tab:food_uncertainty}
    \small

    \resizebox{\textwidth}{!}{%
    \begin{tabular}{llrrrrr|rrr}
        \toprule
        & &
        \multicolumn{5}{c|}{\textbf{Classification over 1,000 repetitions}} &
        \multicolumn{3}{c}{\textbf{Predictive entropy (bits)}} \\
        \textbf{Model} &
        \textbf{Configuration} &
        \textbf{Accuracy} &
        \textbf{MCC} &
        \textbf{Recall} &
        \textbf{Precision} &
        \textbf{FP rate} &
        \textbf{$H_{\mathrm{unsafe}}$} &
        \textbf{$H_{\mathrm{safe}}$} &
        \textbf{Mean $H$} \\
        \midrule

        GPT-5.5 &
        \AMAtabstar &
        \textbf{100.0} & \textbf{1.000} & \textbf{100.0} & \textbf{100.0} & \textbf{0.0} &
        \textbf{0.000} & \textbf{0.000} & \textbf{0.000} \\
        GPT-5.5 &
        RaR &
        96.8 & 0.938 & 93.6 & \textbf{100.0} & \textbf{0.0} &
        0.343 & \textbf{0.000} & 0.172 \\
        GPT-5.5 &
        Basic $\star$ &
        \textbf{100.0} & \textbf{1.000} & \textbf{100.0} & \textbf{100.0} & \textbf{0.0} &
        \textbf{0.000} & \textbf{0.000} & \textbf{0.000} \\
        \midrule

        Claude Opus 4.8 &
        \AMAtabstar &
        \textbf{100.0} & \textbf{1.000} & \textbf{100.0} & \textbf{100.0} & \textbf{0.0} &
        \textbf{0.000} & \textbf{0.000} & \textbf{0.000} \\
        Claude Opus 4.8 &
        RaR &
        99.2 & 0.984 & 98.4 & \textbf{100.0} & \textbf{0.0} &
        0.118 & \textbf{0.000} & 0.059 \\
        Claude Opus 4.8 &
        Basic$^{\dagger}$ &
        99.9 & 0.998 & \textbf{100.0} & 99.8 & 0.2 &
        \textbf{0.000} & 0.021 & 0.010 \\
        \midrule

        Gemini 3.1 Pro &
        \AMAtabstar &
        \textbf{100.0} & \textbf{1.000} & \textbf{100.0} & \textbf{100.0} & \textbf{0.0} &
        \textbf{0.000} & \textbf{0.000} & \textbf{0.000} \\
        Gemini 3.1 Pro &
        RaR &
        99.8 & 0.996 & 99.6 & \textbf{100.0} & \textbf{0.0} &
        0.038 & \textbf{0.000} & 0.019 \\
        Gemini 3.1 Pro &
        Basic $\star$ &
        \textbf{100.0} & \textbf{1.000} & \textbf{100.0} & \textbf{100.0} & \textbf{0.0} &
        \textbf{0.000} & \textbf{0.000} & \textbf{0.000} \\
        \midrule

        LLaMA 3.3 70B &
        \AMAtabstar &
        \textbf{100.0} & \textbf{1.000} & \textbf{100.0} & \textbf{100.0} & \textbf{0.0} &
        \textbf{0.000} & \textbf{0.000} & \textbf{0.000} \\
        LLaMA 3.3 70B &
        RaR $\star$ &
        \textbf{100.0} & \textbf{1.000} & \textbf{100.0} & \textbf{100.0} & \textbf{0.0} &
        \textbf{0.000} & \textbf{0.000} & \textbf{0.000} \\
        LLaMA 3.3 70B &
        Basic $\star$ &
        \textbf{100.0} & \textbf{1.000} & \textbf{100.0} & \textbf{100.0} & \textbf{0.0} &
        \textbf{0.000} & \textbf{0.000} & \textbf{0.000} \\
        \midrule

        Mistral Large &
        \AMAtabstar &
        \textbf{100.0} & \textbf{1.000} & \textbf{100.0} & \textbf{100.0} & \textbf{0.0} &
        \textbf{0.000} & \textbf{0.000} & \textbf{0.000} \\
        Mistral Large &
        RaR $\star$ &
        \textbf{100.0} & \textbf{1.000} & \textbf{100.0} & \textbf{100.0} & \textbf{0.0} &
        \textbf{0.000} & \textbf{0.000} & \textbf{0.000} \\
        Mistral Large &
        Basic $\star$ &
        \textbf{100.0} & \textbf{1.000} & \textbf{100.0} & \textbf{100.0} & \textbf{0.0} &
        \textbf{0.000} & \textbf{0.000} & \textbf{0.000} \\
        \bottomrule
    \end{tabular}%
    }
\end{table}

\FloatBarrier

% #################### DISCUSSION ####################
\section{Discussion}

\subsection{Interpretation of the findings}

The central result is that when a health query is underspecified because
the patient has not disclosed relevant context, acquiring that context is
far more effective than improving the query. Rephrase-and-Respond
performed almost indistinguishably from direct prompting on both tasks:
rewriting can clarify wording, but it cannot recover facts that were
never stated. The \AMA addressed this absence directly, and its gains
were concentrated exactly where the withheld information was decisive.
In diagnosis retrieval, the visible symptoms were usually sufficient to
identify a broad diagnostic family, and clarification added little at
that level; the withheld symptoms mattered for discriminating among
members of the family, which is where exact-diagnosis accuracy rose from
the low single digits to well above half. The small abstention rate is
consistent with this reading: abstention occurred in the deterministic
controller, before any language generation, when the knowledge graph
could not support a single answer or a coherent family. In the dietary
task, where the comparator prompts already contained the patient's
context, the agent still achieved the best MCC for four of five models
and consistently reduced false negatives, suggesting that structured
graph evidence contributed beyond the raw context; the one exception
reflects a different point on the recall--precision trade-off rather
than a failure.

The second finding, that performance after clarification depended far
less on which downstream model was used, has practical implications.
Once the decisive evidence has been moved out of the model's parametric
guesswork and into a clarified prompt, the downstream model is left with
the narrower task of verifying, ranking and formatting, so the choice of
model, and its cost, matters less, and system behaviour becomes more
predictable and more auditable because the questions asked and the
hypotheses eliminated are recorded symbolically.

The uncertainty analysis should be read with two caveats that the
results themselves make clear. Low output entropy is not evidence of
correctness: in the diagnosis task, several baseline configurations returned the same wrong
answer on every repetition, so consistency of a language model's output
is not, by itself, a safety signal. And because the clarification
trajectory was held fixed in the \AMA condition, the zero-entropy result
characterizes how stably a downstream model expresses an already
resolved decision, not the stability of the full interactive process.
What the analysis does show is that once context has been acquired and
encoded as structured evidence, the remaining variation is largely
confined to wording rather than to the clinical conclusion.

Taken together, these findings support a design principle rather than a
particular model: contextual ambiguity is best handled by an intermediary
that decides, from structured domain knowledge, what is missing and what
to ask, while the language model is confined to language-facing
operations. This makes the framework complementary to the remedies for
the other ambiguity types. Retrieval augmentation addresses what the
model does not know, query rewriting addresses what the query does not
say clearly, and the \AMA addresses what the patient has not said at all,
without modifying the downstream model.

\subsection{Future Work}

The clarification variables in this study were answered by the patient,
because the benchmarks were constructed to withhold exactly the
information a patient could report. Nothing in the agent requires
this. A clarification variable is a graph-defined fact whose value the
controller does not yet know, and the controller is indifferent to where
the value comes from. The most immediate extension is therefore to let
the agent resolve variables from data before it resolves them from the
patient. In a deployed setting, symptoms, diagnoses, medications,
allergies, laboratory values and recent measurements are often already
available in an electronic health record, a personal health record or a
connected device, and the openCHA framework on which the agent is built
already provides the orchestration layer for querying such sources.
Under this extension the simulation oracle used here would be replaced
by a set of data connectors, a question would be posed to the patient
only when no connected source could answer it, and the burden of
clarification on the patient would fall accordingly.

A second extension follows from the first: the variables need not be
binary or textual. Graph nodes can represent facts that are best
obtained from other modalities, such as heart-rate or sleep patterns
from a wearable, a photograph of a skin lesion or a swollen joint, or a
voice sample, and the question-selection step can be generalized to
weigh the expected information gain of each variable against the cost
and delay of acquiring it. This cost-aware selection would also make it
natural to prefer a passive data lookup over an intrusive question when
both could resolve the same uncertainty, and to respect the patient's
privacy preferences about which sources may be consulted.

Third, the hypothesis update used here is a hard consistency rule over
binary answers. Real patients answer with uncertainty, sometimes
incorrectly, and sometimes contradict themselves. Replacing the hard
update with a graded, probabilistic update would allow the controller to
tolerate noisy or partial answers, to represent that a symptom is
``sometimes'' present, and to revisit a question when later evidence
casts doubt on an earlier answer. The same relaxation would allow the
agent to maintain more than one hypothesis at the end of clarification,
which is necessary for patients with comorbid conditions.

Finally, because the framework is defined by its knowledge graph and its
output definition rather than by a task-specific controller, it can be
instantiated for other patient-facing tasks in which missing context is
decisive, including medication guidance, where interactions and
contraindications depend on undisclosed prescriptions; symptom triage;
and mental-health support, where the Ambiguity Injection Module has
already been designed to preserve risk-relevant language while masking
detail. The paired records produced by that module can also serve as
supervision for learned clarification policies, so that the deterministic
controller used here can be complemented, where appropriate, by a policy
that has learned which questions patients find easiest to answer.

\subsection{Limitations}

Several limitations bound the interpretation of these results. The
diagnosis benchmark was derived from synthetic patients generated by
Synthea, and the dietary benchmark from authored query templates; neither
contains real patient language, and the diagnosis label distribution was
concentrated, with three respiratory diagnoses accounting for more than
two thirds of the evaluation set. Performance on naturally occurring
queries, with their misspellings, colloquial symptom descriptions and
mixed concerns, remains to be established.

The clarification answers were supplied by a simulation oracle that
always answered correctly and in binary form. Real patients may be
uncertain, may misreport, or may answer a different question from the one
asked, and the hard elimination rule used here is sensitive to a single
incorrect answer. The interpretation of free-text replies as binary
answers was likewise not stressed in this evaluation. The interaction cost
of clarification was also not assessed from the patient's side: the agent
asked up to five questions in the diagnosis task and a mean of 5.96, and
at most 11, in the dietary task, and whether patients tolerate this
number of questions in practice is a question for user studies.

The agent inherits the coverage and accuracy of its knowledge
graphs. A missing entity or relation prevents a hypothesis or question
from being considered at all, and a guideline-derived relation may be too
general for an individual whose portion size, preparation method, disease
severity or comorbidities change its applicability. The
\texttt{HAS\_SYMPTOM} relations used for diagnosis were drawn from UMLS
and were not curated for prevalence or specificity, and the category and
ICD mappings used for the secondary diagnosis outcomes were checked by the
authors but not independently adjudicated by clinicians.

Finally, the study involved no patients, clinicians or deployment. It
establishes that acquiring missing context improves the accuracy and
consistency of downstream responses under controlled conditions; it does
not establish clinical safety, effectiveness, fairness across patient
groups, or usability, and the \AMA should not be understood as an
autonomous diagnostic or dietary decision system. Addressing these
limitations, through evaluation on independently reviewed cases, noisy
and contradictory answers, naturally occurring queries and, ultimately,
prospective use with patients, is the necessary next step before an
agent of this kind can be placed in front of real users.
% ================================================================
% METHODS
% ================================================================

\section{Methods}
\label{sec:methods}

This section has two parts. The first part describes the proposed method,
the \AMA. It begins with an overview of what the agent is and how it
operates, and then presents each of its components in turn. For each
component, we first describe how it operates in general, independently of
any particular language model or database, and then describe the
implementation used in this study. The second part describes the
evaluation method: the evaluation design, the two benchmarks, the
downstream models and comparator configurations, and the statistical
analysis.

\subsection{Ambiguity Mitigation Agent}

\begin{figure}[!t]
    \centering
    \includegraphics[width=\textwidth]{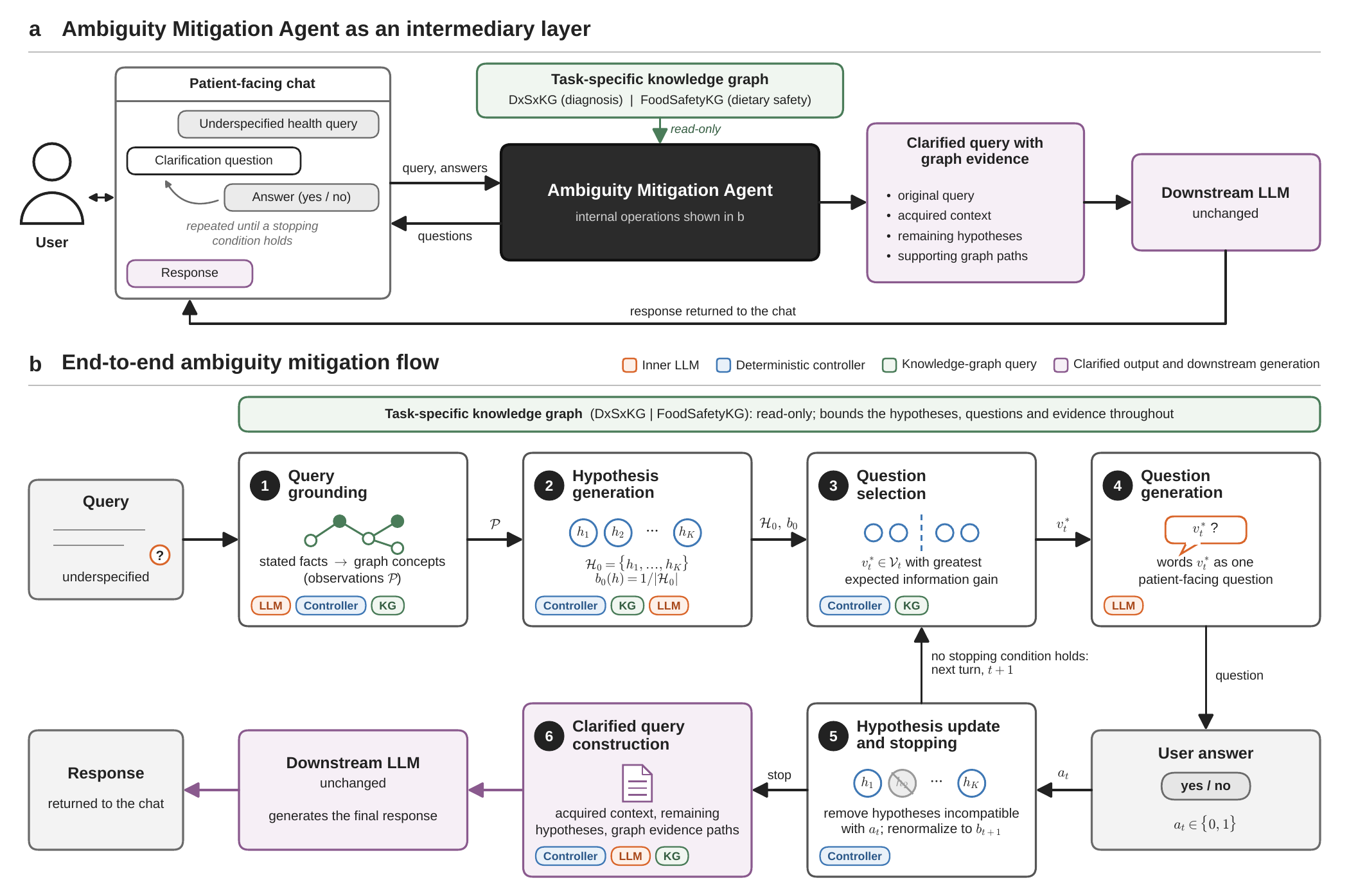}
    \caption{\textbf{The \AMA.}
\textbf{a}, The agent as an intermediary layer. The agent operates between
the patient-facing chat and an otherwise unchanged downstream LLM, and
reads from a task-specific knowledge graph (DxSxKG for diagnosis,
FoodSafetyKG for dietary safety). The chat window shows the exchange: the
user's underspecified query, the agent's clarification questions and the
user's answers, repeated until a stopping condition holds. The agent's
product is a clarified query with graph evidence (the original query, the
acquired context, the remaining hypotheses and the supporting graph paths),
which is the input to the downstream LLM; the response is returned to the
chat.
\textbf{b}, End-to-end ambiguity mitigation flow, from the user's query to
the final response; operations 1--6 take place inside the agent shown in
\textbf{a}. (1) Query grounding maps the query to graph concepts
(observations \(\mathcal{P}\)); (2) hypothesis generation retrieves the
graph-supported hypotheses \(\mathcal{H}_0=\{h_1,\ldots,h_K\}\) and assigns
uniform weights \(b_0\); (3) question selection chooses the clarification
variable \(v_t^{*}\in\mathcal{V}_t\) with the greatest expected information
gain; (4) question generation words it for the user, who replies with a
binary answer \(a_t\in\{0,1\}\); (5) hypothesis update and stopping removes
the hypotheses incompatible with \(a_t\), renormalizes the weights to
\(b_{t+1}\) and tests the stopping conditions, returning to question
selection for the next turn if none holds; (6) clarified query construction
assembles the acquired context, the remaining hypotheses and the graph
evidence paths for the downstream LLM, which generates the final response.
Tags indicate which actor performs each operation. The deterministic
controller (blue) is the rule-based, non-LLM component that makes every
decision that changes the agent's state: which graph entities are accepted,
which hypotheses are retrieved, which question is asked, how an answer
updates the hypotheses and when clarification stops. The inner LLM (orange)
performs only the language-facing steps, and knowledge-graph queries are
shown in green; purple denotes the clarified output and downstream
generation. The knowledge graph is read-only and bounds the hypotheses,
questions and evidence throughout.}
\label{fig:framework}
\end{figure}

The \AMA is an intermediary layer between a user and an otherwise unchanged
downstream large language model (LLM) (Fig.~\ref{fig:framework}a). It receives
an underspecified health query, determines which patient-specific facts are
missing, obtains them by asking the user targeted questions, and returns a
clarified query together with the supporting evidence to the downstream LLM,
which generates the final response. The agent does not generate the final
response itself and does not modify the downstream model.

The agent treats clarification as hypothesis elimination over a task-specific
knowledge graph. Let \(\mathcal{P}\) denote the set of observations, that is,
the graph concepts grounded from the facts stated in the query. The graph
defines a set of hypotheses \(\mathcal{H}\), the candidate answers that the
observations could support (candidate diagnoses in the diagnosis task,
food-associated risk conditions in the dietary-safety task), and a set of
binary clarification variables \(\mathcal{V}\), patient-reportable facts
connected to hypotheses by support relations (symptoms in the diagnosis task,
health conditions and condition families in the dietary-safety task). Because
each variable is connected to some hypotheses and not to others, the answer to
a variable partitions the active hypothesis set. The agent therefore proceeds
in a loop (Fig.~\ref{fig:framework}b): it grounds the query in the graph,
retrieves the supported hypotheses, selects the variable whose answer is
expected to reduce uncertainty over the hypotheses the most, asks it, removes
the hypotheses that are incompatible with the answer, and repeats until a
stopping condition holds. It then assembles the acquired context and the
graph evidence into a clarified query.

Three actors carry out these steps. A deterministic controller makes every
decision that changes the agent's state: which graph entities are accepted,
which hypotheses are retrieved, which question is asked, how an answer updates
the hypotheses and when clarification stops. An inner LLM performs the
language-facing operations: interpreting the query, wording the selected
variable as a question and formatting the clarified record. The inner LLM
proposes and formats but cannot alter the controller's state, so the
clarification trajectory is reproducible given the same answers. The knowledge
graph bounds what the agent can consider: only graph-supported hypotheses can
be retrieved, only graph-defined variables can be asked and only graph paths
can be passed on as evidence. Domain knowledge is thus encoded in the graph,
whereas the clarification logic is identical across tasks; instantiating the
agent for a new task requires a new graph and a task-specific output
definition, not a new controller.

The six operations in Fig.~\ref{fig:framework}b are described in the
following sections in the order in which they occur: query grounding,
hypothesis generation, question selection, question generation, hypothesis
update and stopping, and clarified query construction. The knowledge graph,
which all of them consult, is described first.

The agent was implemented in Python within
openCHA~\cite{abbasian2025conversational}. The inner LLM was GPT-5.5, accessed
through the OpenAI API and held fixed in every experiment; only the downstream
response model varied across the evaluated model panel. The task-specific
knowledge graphs were stored in Neo4j and accessed through the official Neo4j
Python driver using read-only Cypher queries; Neo4j was an implementation
choice rather than a requirement of the method. In interactive use, questions
and answers pass through a chat interface built with Gradio; in the benchmark
runs described under ``Evaluation method'', answers were supplied
programmatically by a simulation oracle. Two instantiations of the agent were
evaluated, one for symptom-based diagnosis retrieval and one for dietary-safety
classification; Table~\ref{tab:instantiation} summarizes how each component
was instantiated in the two tasks. No LLM was trained or fine-tuned. Verbatim
inner-model prompts and output schemas are given in the Supplementary
Information.

\begin{table}[t]
    \centering
    \caption{\textbf{Instantiation of the \AMA components in the two
    evaluated tasks.} The controller logic is shared; the task determines
    the knowledge graph, the meaning of observations, hypotheses and
    clarification variables, the stopping rules and the output passed to the
    downstream LLM.}
    \label{tab:instantiation}
    \small
    \begin{tabularx}{\textwidth}{
        >{\raggedright\arraybackslash}p{0.19\textwidth}
        >{\raggedright\arraybackslash}X
        >{\raggedright\arraybackslash}X}
        \toprule
        \textbf{Component} &
        \textbf{Symptom-based diagnosis retrieval} &
        \textbf{Dietary safety classification} \\
        \midrule
        Knowledge graph &
        DxSxKG: diagnoses and symptoms from UMLS; \texttt{HAS\_SYMPTOM} support
        relations; \texttt{IS\_A} hierarchy &
        FoodSafetyKG: foods, ingredients, nutrients, food properties and
        patient conditions from USDA FoodData Central, UMLS and clinical
        guidelines; \texttt{RISKY\_FOR} and \texttt{SAFE\_FOR} relations;
        condition families \\
        Observations \(\mathcal{P}\) &
        Visible symptom mentions resolved to symptom concepts &
        Food terms resolved to food phrases or items \\
        Hypotheses \(\mathcal{H}\) &
        Candidate diagnoses connected to the observations (at most 8) &
        Patient conditions carrying a graph-supported food risk (at most 50) \\
        Clarification variables \(\mathcal{V}\) &
        Unasked symptoms connected to active diagnoses &
        Individual conditions and condition families \\
        Stopping conditions &
        Leading posterior \(>0.95\); no question with positive gain; five
        questions asked &
        Risk condition confirmed or all denied; leading posterior
        \(>0.95\); no question with positive gain; no fixed turn limit \\
        Terminal states &
        \textsc{Answer}, \textsc{Partial} (shared \texttt{IS\_A} family),
        \textsc{Abstain} &
        \texttt{NOT\_OKAY} hint, \texttt{OKAY} hint, or no hint when
        clarification is incomplete \\
        Downstream output &
        Up to five ranked diagnoses &
        One binary safety decision \\
        \bottomrule
    \end{tabularx}
\end{table}

\subsubsection{Knowledge graph}

The knowledge graph is the agent's only source of domain knowledge and is read
but never written during a conversation. It contains two kinds of nodes that
the controller reasons over, hypothesis nodes and clarification-variable
nodes, connected by support relations that state which variables are expected
to be present under which hypotheses. Hierarchy relations group hypotheses
into families, which allows the agent to return a clinically coherent group
when the evidence cannot single out one member. Alias and synonym entries,
together with a text index over node names, support the grounding of free-text
mentions. From these relations the controller derives everything it needs: the
hypotheses supported by a set of observations, the variables that can still
distinguish the active hypotheses, the partition of the active set induced by
each possible answer, and the evidence paths that link the acquired context to
the remaining hypotheses. The graph contains no procedural logic and does not
generate any part of the final response. It bounds the hypotheses the agent
can consider, the questions it can ask and the evidence it can pass on.

The symptom--diagnosis knowledge graph (DxSxKG) represented diagnoses and
symptoms as concepts from the 2025AA release of the Unified Medical Language
System (UMLS)~\cite{bodenreider2004unified}. Diagnosis nodes were connected to
symptom nodes by \texttt{HAS\_SYMPTOM} relations and to clinical ancestors by
\texttt{IS\_A} relations. The evaluated graph contained 129,815 diagnosis
nodes, 340,842 symptom nodes, 82,813 \texttt{HAS\_SYMPTOM} relations and 1,400
\texttt{IS\_A} relations. The relations used by the agent were obtained from
UMLS and normalized to the schema described in Supplementary Methods. In this
instantiation, diagnoses are the hypotheses, symptoms are the clarification
variables, \texttt{HAS\_SYMPTOM} defines both candidate support and the
partition associated with each question, and \texttt{IS\_A} is used only to
identify a shared diagnostic family when the evidence does not resolve a
single diagnosis. The held-out test partition of the diagnosis benchmark did
not contribute relations to DxSxKG.

FoodSafetyKG integrated the USDA FoodData Central full CSV release dated
18 December 2025~\cite{usda_fdc}, the 2025AA release of the UMLS
Metathesaurus~\cite{bodenreider2004unified,nlmUMLS2025AA}, and a registry of
clinical-guideline-derived rules spanning 14 guideline families covering areas
including diabetes, cardiovascular disease, gastrointestinal disorders, kidney
disease, pregnancy, food allergy and other dietary-risk contexts. FoodData
Central provided food-composition and nutrient records. UMLS provided
normalized condition concepts, aliases and parent--child relations. The
guideline registry encoded associations between food properties and risk or
safety for specific patient conditions, with the source and evidence type
retained for each relation. The evaluated graph contained 2,085,340 food-item
nodes, 477 nutrient nodes, 27,094,028 \texttt{HAS\_NUTRIENT} relations,
2,542,284 derived \texttt{HAS\_PROPERTY} relations and 21,030,444 materialized
food-item risk relations. It also represented food phrases, ingredients, food
properties, patient conditions, condition families and aliases. Food phrases
and food items were connected to ingredients, nutrients and food properties,
which were linked to conditions by guideline-supported \texttt{RISKY\_FOR} and
\texttt{SAFE\_FOR} relations. In this instantiation, patient conditions are the
hypotheses, conditions and condition families are the clarification
variables, \texttt{RISKY\_FOR} defines support and \texttt{SAFE\_FOR} encodes
exceptions. The complete graph was available throughout the evaluation;
benchmark queries were resolved through the curated food-expression layer,
comprising 298 curated food expressions linked to ingredients and 60
nutritional or biochemical food properties, rather than through the complete
USDA food-item layer. Construction and provenance details for both graphs are
provided in Supplementary Methods.

\subsubsection{Query grounding}

Query grounding maps the free-text query to the observation set
\(\mathcal{P}\), the graph concepts corresponding to the facts the user has
stated. The inner LLM identifies the spans of the query that require grounding
and proposes search terms, marking any mention whose wording could support
more than one clinical interpretation. The controller then resolves each term
against the graph vocabulary in a fixed order, normalized concept names,
stored synonyms and finally the full-text index, and retains a bounded number
of matches: several for a short or ambiguity-flagged mention and the single
highest-ranked match otherwise. If these searches return nothing, the inner
LLM may propose canonical vocabulary wording, which the controller accepts
only when it matches an eligible graph node exactly. Terms that cannot be
resolved are excluded rather than guessed. The inner LLM therefore proposes
and the controller disposes: nothing enters \(\mathcal{P}\) that does not
exist in the graph, and neither the reference answer nor any withheld
information is available at this stage.

In the diagnosis task, the visible symptom mentions were grounded to DxSxKG
symptom concepts. The resolver retained up to three graph matches for a short
or ambiguity-flagged mention and the highest-ranked match for any other
mention. In the dietary-safety task, the inner model identified one or more
food terms from the visible text, and a deterministic extractor supplied a
fallback term when needed; each term was resolved to a FoodSafetyKG food
phrase or food item. The grounding prompts are given in Supplementary
Information.

\subsubsection{Hypothesis generation}

Given \(\mathcal{P}\), the controller queries the graph for the hypotheses
connected to the observations, ranks them and truncates the list at a
task-specific maximum. Before the hypothesis set is fixed, the inner LLM makes
one bounded expansion decision: whether a clinically plausible interpretation
of the visible query is absent from the retrieved hypotheses. If so, it may
propose up to three additional search terms; a proposed term affects retrieval
only when it resolves to a graph entity and, where the task requires it,
returns at least one graph-supported hypothesis. Expansion is performed once,
before clarification begins, and without access to withheld information. The
resulting ranked list forms the initial hypothesis set
\(\mathcal{H}_0=\{h_1,\ldots,h_K\}\), and the controller assigns each
hypothesis equal initial weight,
\begin{equation}
 b_0(h)=\frac{1}{|\mathcal{H}_0|}, \qquad h\in\mathcal{H}_0.
 \label{eq:uniform-prior}
\end{equation}
These weights guide question selection and are not interpreted as calibrated
clinical probabilities. If no graph-supported hypothesis is retrieved, the
controller enters the unresolved state defined for the task rather than
substituting a model-generated hypothesis.

In the diagnosis task, let \(\mathcal{P}\) denote the graph concepts obtained
from the visible query and the single expansion step. When
\(|\mathcal{P}|=1\), a diagnosis was eligible if it was connected to that
concept; otherwise, it had to be connected to at least two concepts in
\(\mathcal{P}\). The graph query returned at most eight diagnoses,
prioritizing greater concept overlap, greater proportional coverage of
\(\mathcal{P}\) and fewer graph-connected symptoms. Fixed diagnosis-name
exclusions were then applied without replacing excluded candidates. Each
remaining candidate \(h\) was ranked by
\begin{equation}
 s(h)=2k_h+\frac{k_h}{m}-0.01n_h,
 \label{eq:dx-retrieval-score}
\end{equation}
where \(k_h\) is the number of concepts in \(\mathcal{P}\) matched by \(h\),
\(m=|\mathcal{P}|\), and \(n_h\) is the number of symptoms connected to \(h\)
in DxSxKG. The first two terms reward direct and proportional overlap with the
mapped query, whereas the final term gives a small preference to more specific
diagnoses. Graph-matched expansion terms were used together with the
query-derived concepts to retrieve the final candidates and were not asked
again as clarification questions.

In the dietary-safety task, the controller used each grounded food term to
search FoodSafetyKG. It first retrieved conditions connected by direct
\texttt{RISKY\_FOR} relations and, when no direct relation was found,
traversed the permitted food--ingredient--property--condition path. Conditions
returned for multiple food terms were combined and deduplicated, so that a
patient condition could enter the hypothesis set only through graph-supported
food-risk evidence. A proposed expansion term affected retrieval only when it
matched FoodSafetyKG and returned at least one graph-supported risk condition.
The union of the retrieved conditions was limited to 50 and formed
\(\mathcal{H}_0\).

\subsubsection{Question selection}

At turn \(t\), let \(\mathcal{H}_t\) be the active hypotheses with normalized
weights \(b_t\), and let \(\mathcal{V}_t\) be the eligible clarification
variables: the previously unasked variables connected to at least one active
hypothesis, excluding those already stated in the query. For a variable
\(v\) and an answer \(a\in\{0,1\}\), \(\mathcal{H}_t(v,a)\subseteq\mathcal{H}_t\)
denotes the hypotheses compatible with that answer. Uncertainty over the
hypotheses is measured by the Shannon entropy of the weight distribution,
\begin{equation}
 H(b_t)=-\sum_{h\in\mathcal{H}_t}b_t(h)\log_2 b_t(h).
 \label{eq:belief-entropy}
\end{equation}
The controller selects the variable with the greatest expected reduction in
entropy,
\begin{equation}
 v_t^*=\arg\max_{v\in\mathcal{V}_t}
 \left\{
 H(b_t)-\sum_{a\in\{0,1\}}P(a\mid v,b_t)
 H\!\left(b_t\mid a,v\right)
 \right\},
 \label{eq:information-gain}
\end{equation}
where \(P(a\mid v,b_t)=\sum_{h\in\mathcal{H}_t(v,a)}b_t(h)\), and
\(H(b_t\mid a,v)\) is the entropy after restriction to
\(\mathcal{H}_t(v,a)\) and renormalization. Only variables with positive
expected gain are considered, and exact ties are resolved deterministically
without access to the withheld reference answer. Because the selection depends
only on the graph relations and the current weights, the same variable is
selected regardless of which inner or downstream LLM is used.

In the diagnosis task, the eligible variables were the unasked symptoms
connected to diagnoses that still had non-zero weight. Symptoms already
represented in the query, previously answered or removed by a fixed question
filter were not eligible. In the dietary-safety task, two question types were
available: family questions, defined by the condition-family relations in
FoodSafetyKG, which partition several hypotheses at once, and leaf questions
about individual conditions. At each turn the controller considered both types
and selected the partition with the greatest expected information gain. When
only one unasked condition remained, the controller asked about it directly
rather than inferring that it was present.

\subsubsection{Question generation}

The inner LLM renders the selected variable \(v_t^*\) as a single
patient-facing question in natural language. The graph-defined meaning of the
variable is fixed: the inner LLM chooses the wording but cannot change which
symptom, condition or family is being asked about, and it cannot introduce a
question that the controller did not select. Questions are never repeated. The
user's reply is interpreted as a binary answer \(a_t\in\{0,1\}\) to
\(v_t^*\) and returned to the controller.

Question wording used a fixed prompt and a structured output schema with
validation and retry rules (Supplementary Information). In the diagnosis task,
a ``yes'' answer meant that the symptom was present. In the dietary-safety
task, a ``yes'' answer to a family question meant that the patient's condition
belonged to that family, and a ``yes'' answer to a leaf question meant that
the patient had that condition. In interactive use the answer is typed by the
user; in the benchmark runs it was supplied by the simulation oracle described
under ``Evaluation method''.

\subsubsection{Hypothesis update and stopping}

After the answer \(a_t\), the controller removes the hypotheses that are
incompatible with it and renormalizes the remaining weights,
\begin{equation}
 b_{t+1}(h)=
 \frac{\mathbf{1}[h\in\mathcal{H}_t(v_t^*,a_t)]b_t(h)}
 {\sum_{h'\in\mathcal{H}_t}
  \mathbf{1}[h'\in\mathcal{H}_t(v_t^*,a_t)]b_t(h')}.
 \label{eq:belief-update}
\end{equation}
If no hypothesis remains, the controller records an unresolved state rather
than substituting a model-generated hypothesis. The controller then evaluates
the stopping conditions. Clarification stops when the remaining hypotheses
support a task output, that is, when the leading hypothesis exceeds a
posterior threshold or a task-specific confirmation is reached; when no
unasked variable has positive expected gain, so that no further question could
distinguish the remaining hypotheses; or when the task's interaction limit is
reached. With uniform initial weights and the hard consistency update above,
the posterior threshold is reached exactly when a single hypothesis remains.
The terminal state records whether the evidence supports a single hypothesis
(\textsc{Answer}), a clinically coherent family of hypotheses
(\textsc{Partial}), or neither (\textsc{Abstain} or unresolved). If no
stopping condition holds, control returns to question selection.

In the diagnosis task, a ``yes'' answer removed the diagnoses not connected to
the asked symptom, whereas a ``no'' answer removed the diagnoses that were
connected to it. Clarification ended when no initial diagnosis had been
retrieved, the leading diagnosis had posterior probability greater than 0.95,
no eligible symptom had positive information gain, or five questions had been
asked. A single remaining diagnosis produced \textsc{Answer}. If the
controller instead stopped because no informative question remained or the
question limit was reached, it searched for an eligible immediate
\texttt{IS\_A} ancestor shared by at least \(\lceil|\mathcal{H}_0|/2\rceil\)
of the initially retrieved diagnoses; a qualifying family produced
\textsc{Partial}, and otherwise the controller returned \textsc{Abstain}.
Failure to retrieve an initial diagnosis also produced \textsc{Abstain}. A
diagnosis generated after abstention was retained only for counterfactual
inspection and was not scored as an answered case.

In the dietary-safety task, a ``yes'' answer to a family question retained the
conditions belonging to that family, whereas a ``no'' answer removed them; an
answer to an individual-condition question retained or removed that condition
in the same way. Clarification ended when a food-associated risk condition was
confirmed, all graph-supported risk conditions were denied, the leading
hypothesis exceeded posterior probability 0.95, or no remaining question had
positive information gain. Failure to retrieve an initial risk hypothesis
ended clarification with limited evidence. The food-safety controller had no
fixed numerical turn limit; the finite graph-defined question set bounded the
interaction.

\subsubsection{Clarified query construction}

When clarification ends, the controller fixes the clarified record: the
original query, the question--answer history, the remaining hypotheses and
their weights, the resolution state, the stopping reason and the graph
evidence paths that link the acquired context to the remaining hypotheses.
The inner LLM formats this record as a structured clarified query for the
downstream LLM; it may organize the information but cannot change the
selected questions, the hypothesis updates, the stopping decision or any
symbolic conclusion recorded by the controller. The downstream LLM receives
the clarified query, the structured graph evidence and the task-specific
output instruction, and generates the response. The clarified query is the
only channel through which the agent influences the downstream model, which
is otherwise unchanged.

In the diagnosis task, the clarified record contained the ranked remaining
diagnoses and the terminal state, and the downstream LLM returned up to five
ranked diagnoses. In the dietary-safety task, the controller retrieved the
graph paths supporting the confirmed or remaining risk conditions and checked
whether a confirmed condition also had an applicable \texttt{SAFE\_FOR}
relation. Confirmation of a risk condition without a safety exception produced
a symbolic \texttt{NOT\_OKAY} hint, whereas denial of all candidate risks or
an applicable safety exception produced an \texttt{OKAY} hint; no symbolic
class was assigned when graph-supported clarification remained incomplete.
The inner model assembled the original query, the disclosed answers, the
remaining conditions, the stopping reason and the graph evidence into the
clarified record, and the downstream LLM returned one binary prediction.
Clarification completeness was recorded separately and was not treated as a
third prediction class. The downstream prompts and output schemas for both
tasks are given in Supplementary Information.

\subsection{Evaluation method}

\subsubsection{Evaluation design}
\label{sec:study-design}

The \AMA was evaluated in two health-related tasks that require information
absent from the initial query and that differ in output structure:
symptom-based diagnosis retrieval from incomplete symptom profiles and
dietary-safety classification without the patient's health condition. Both
evaluation sets were constructed so that the intended answer could not be
uniquely determined from the information initially presented to the model.
Each benchmark instance therefore paired an ambiguous query with a complete
reference case; the missing patient-specific information was retained
separately and could be recovered only through the clarification process. In
the \AMA condition, the controller received only the ambiguous query, and a
simulation oracle accessed the withheld field solely to answer the questions
selected by the controller. In a live interaction, those answers would be
provided by the user. Table~\ref{tab:task_overview} summarizes the two
evaluation sets; their construction is described under the task-specific
headings below.

Three configurations were evaluated with five downstream language models
(GPT-5.5, Claude Opus 4.8, Gemini 3.1 Pro, LLaMA 3.3 70B and Mistral Large;
see ``Downstream models and comparator configurations''). In the
\textsc{Basic} configuration, the downstream model answered the supplied query
directly without clarification. In Rephrase-and-Respond
(\textsc{RaR})~\cite{Deng2024}, the model first reformulated the supplied
query but acquired no new patient-specific information. In the \AMA
configuration, targeted follow-up questions were used to acquire the withheld
patient context before the clarified query was passed to the same downstream
model. No LLM was trained or fine-tuned for this study.

The unit of analysis was the case--query instance. The diagnosis benchmark
contained 1,034 instances and the food-safety benchmark contained 487. The
primary diagnosis outcome was exact agreement at rank 1, and the primary
food-safety outcome was the Matthews correlation coefficient (MCC). The
performance of the \AMA was assessed along three dimensions: the accuracy of
the downstream response, the consistency of that performance across the five
downstream models, and the uncertainty of the downstream output under
repeated generation (see ``Statistical analysis''). All reported analyses were
calculated from frozen row-level outputs.

\begin{table}[t]
    \centering
    \caption{\textbf{Overview of the two ambiguity-mitigation evaluation sets.}
    Both evaluation sets were curated so that the intended output could
    not be uniquely determined from the information initially provided.
    Missing patient-specific information was subsequently made available
    through the clarification process.}
    \label{tab:task_overview}
    \small
    \begin{tabularx}{\textwidth}{
        >{\raggedright\arraybackslash}p{0.27\textwidth}
        >{\raggedright\arraybackslash}X
        >{\raggedright\arraybackslash}X}
        \toprule
        &
        \textbf{Symptom-based diagnosis retrieval} &
        \textbf{Dietary safety classification} \\
        \midrule

        Data source &
        Curated synthetic patient records &
        Curated food and health-condition resources \\

        Evaluation instances &
        1,034 &
        487 \\

        Output space &
        32 diagnosis labels &
        \textsc{OK} or \textsc{Not OK} \\

        Information initially provided &
        Partial symptom profile &
        Food-related question without complete health context \\

        Information initially withheld &
        Approximately 49.4\% of the symptoms recorded in the complete case &
        Patient-specific health conditions relevant to the queried food \\

        Information initially visible &
        Approximately 50.6\% of the symptoms recorded in the complete case &
        Queried food and the user's stated request \\

        Reference information used during clarification &
        Complete symptom profile &
        Complete patient health context \\

        Health conditions represented &
        -- &
        38 \\

        Label distribution &
        32 diagnosis classes &
        301 \textsc{OK} and 186 \textsc{Not OK} \\

        Downstream models &
        GPT-5.5, Claude Opus 4.8, Gemini 3.1 Pro, LLaMA 3.3 70B, Mistral Large &
        GPT-5.5, Claude Opus 4.8, Gemini 3.1 Pro, LLaMA 3.3 70B, Mistral Large \\

        Configurations compared &
        \textsc{Basic}, \textsc{RaR}, \AMA &
        \textsc{Basic}, \textsc{RaR}, \AMA \\

        Primary evaluation measures &
        Category-level, ICD-level, and exact diagnosis retrieval
        (Top-1 accuracy and Recall@5) &
        MCC, recall, precision, and classification error rates \\

        \bottomrule
    \end{tabularx}
\end{table}

\subsubsection{Symptom--diagnosis benchmark}
\label{sec:dx-task}

The diagnosis task measured retrieval of a reference diagnosis from incomplete
synthetic evidence; it did not evaluate autonomous diagnosis, triage or
patient care. Each instance contained a reference diagnosis, a complete
symptom profile and a partition of the symptoms into visible and withheld
sets. In every configuration, the downstream model was asked for up to five
ranked diagnoses; in the \AMA configuration, the agent received only the
visible symptoms.

\paragraph{Dataset and cohort construction.}

The benchmark was derived from Synthea, an open-source patient simulator that
produces realistic longitudinal health records from clinical and
epidemiological models of population-level disease
patterns~\cite{walonoski2018synthea}. The source cohort was generated on
24 August 2025 (UTC) from Synthea commit \texttt{c807432}
(\texttt{master-branch-latest}), with the CSV and symptom exporters enabled.
The run simulated Florida residents aged 0--90 years using all disease
modules, a 10-year history, generation seed 404 and reference date
24 August 2025. The run produced 4,884 synthetic patients.

Benchmark construction selected disorder-labelled records with recorded
symptoms and retained 40 reference pathologies. For each eligible case, the
reference diagnosis and the complete associated symptom set were retained from
the original synthetic record without manually changing the diagnosis or
rewriting the symptom content. Patients without a retained case were excluded,
leaving 4,518 patients and 10,374 case--diagnosis instances. Patients, rather
than individual case rows, were assigned to the training, validation and test
partitions (3,163, 904 and 451 patients, respectively), so that a patient
could not occur in more than one partition. Controlled ambiguity was then
introduced for each instance by randomly withholding a non-empty subset of the
complete symptom profile while at least one symptom remained visible. Cases
were retained only when the remaining visible symptoms were compatible with
multiple candidate diagnoses, ensuring that the initial query was genuinely
underspecified rather than simply shorter. The withheld set was stored
separately and was available only to the simulation oracle. Complete
provenance, partition counts and file hashes are provided in the Supplementary
Methods.

The fixed evaluation cohort comprised 1,034 case--diagnosis instances from 451
synthetic patients and covered 32 reference diagnoses. The complete cases
contained 8,503 symptom occurrences (mean 8.22 symptoms per case): 4,299
(50.6\%; mean 4.16 per case) were initially visible and 4,204 (49.4\%; mean
4.07 per case) were withheld. Patient identifiers did not overlap across the
supplied training, validation and test partitions. No LLM was fitted on these
partitions, and the held-out test partition was not used to construct the
knowledge graph.

\paragraph{Simulated user.}

Clarification answers were supplied by a simulation oracle with access to the
complete symptom profile. When the agent asked about a symptom, the oracle
returned ``yes'' if that symptom occurred in the complete profile and ``no''
otherwise. The oracle answered only the questions selected by the controller
and exposed no other part of the withheld information.

\paragraph{Outcome measures.}

Diagnosis retrieval was evaluated using Top-1 accuracy and Recall@5 at three
levels of specificity: exact diagnosis name, ICD-level grouping and broad
diagnostic category. The primary outcome was exact agreement between the
reference diagnosis and the first returned diagnosis (exact Top-1). Recall@5
recorded whether the reference diagnosis appeared among the up to five
returned diagnoses. Exact agreement required case-insensitive equality after
trimming and normalizing whitespace. For the secondary concept-based analyses,
returned and reference names were resolved to UMLS concepts and assigned the
ICD-10 mappings associated with those concepts. Broader diagnostic categories
were derived from the corresponding UMLS hierarchy. The authors checked the
final mapping table before calculating these outcomes; the category and ICD
mappings were not independently clinician-adjudicated.

Selective outcomes used answered cases as the denominator and were accompanied
by coverage. Overall outcomes used all 1,034 instances and scored abstentions
as incorrect. Because the \textsc{Basic} and \textsc{RaR} configurations did
not abstain, their selective and overall values coincide. Clarification
outcomes included the initial and terminal hypothesis-set sizes, entropy
reduction, number of questions, stopping reason and terminal-state frequency.

\subsubsection{Dietary-safety benchmark}
\label{sec:food-task-rewrite}

The dietary-safety task tested whether the agent could recover patient context
omitted from a dietary question before a downstream LLM classified the
request. Each instance paired an ambiguous food query with a withheld patient
condition and a binary reference decision, \texttt{OKAY} or
\texttt{NOT\_OKAY}. In the \AMA configuration, the agent initially received
only the ambiguous query; the withheld condition was available only to the
simulation oracle.

\paragraph{Dataset construction.}

The food-safety benchmark comprised 487 instances spanning 38 patient health
contexts: 301 labelled \texttt{OKAY} and 186 labelled \texttt{NOT\_OKAY}.
Each instance paired a food-related query with a patient-specific context and
a binary reference decision, and also retained the corresponding fully
specified query and reference explanation. The Ambiguity Injection Module
created the ambiguous query by removing the phrase that specified the
patient's health context while leaving the food request and reference
decision unchanged, so that the safety decision could not be resolved from
the visible food description alone. The removed context was stored separately
for oracle simulation. The withheld information represented clinically
relevant factors including metabolic conditions such as diabetes and
hypertension, gastrointestinal disorders such as GERD and IBS, food allergies
and intolerances, dietary restrictions and pregnancy-related context.

\paragraph{Simulated user.}

The simulation oracle used the withheld patient condition to answer the
agent's questions: ``yes'' to a family question when the withheld condition
belonged to that family and ``yes'' to a leaf question when it was that
condition, and ``no'' otherwise. As in the diagnosis task, the oracle answered
only the questions selected by the controller.

\paragraph{Outcome measures.}

MCC was the primary discrimination measure because the dataset was imbalanced
(301 \texttt{OKAY} and 186 \texttt{NOT\_OKAY}), so that accuracy alone could
mask poor performance on the less frequent class, whereas MCC incorporates all
four cells of the confusion matrix. Accuracy, precision, recall and \(F_1\)
were also calculated with \texttt{NOT\_OKAY} as the positive class, together
with the false-negative and false-positive rates; with this convention, recall
and the false-negative rate reflect the system's ability to identify foods
associated with a relevant health risk, and a false positive denotes an
unnecessary restriction. The raw tokens \texttt{OK} and \texttt{NOT\_OK} were
normalized to \texttt{OKAY} and \texttt{NOT\_OKAY}, respectively. A missing or
invalid output was not assigned a class and was excluded from metrics
requiring a valid binary prediction; the corresponding denominator was
reported.

Clarification outcomes included the initial and terminal hypothesis-set sizes,
entropy reduction, number and type of questions, stopping reason and resolution
completeness. All valid binary outputs entered the primary classification
analysis irrespective of whether clarification was complete.

\subsubsection{Downstream models and comparator configurations}
\label{sec:models-comparators}

The downstream panel comprised GPT-5.5 (\texttt{gpt-5.5}; OpenAI), Claude Opus
4.8 (\texttt{claude-opus-4-8}; Anthropic), Gemini 3.1 Pro Preview
(\texttt{google/gemini-3.1-pro-preview}; a Google model served through
OpenRouter), Llama 3.3 70B Instruct
(\texttt{meta-llama/llama-3.3-70b-instruct}; OpenRouter) and Mistral Large
(\texttt{mistral-large-2512}; Mistral AI). The requested identifiers,
provider-returned identifiers when available and access timestamps were stored
in the run manifests. GPT-5.5 was fixed as the inner clarification model; only
the downstream response model varied across the model panel.

Direct prompting (\textsc{Basic}) combined the supplied query with the
task-specific output instruction. RaR used two calls~\cite{Deng2024}: the
first reformulated the supplied query without adding information, and the
second generated a response from the original and reformulated queries. RaR
could not ask clarification questions or access the simulation oracle. In the
\AMA condition, clarification preceded response generation and the downstream
LLM received the clarified query, structured graph evidence and the
task-specific output instruction.

For the diagnosis task, all three configurations began with the same incomplete
symptom presentation. In the frozen food-safety evaluation, the \AMA began
with the ambiguous food query and recovered patient context through simulated
answers, whereas the direct and RaR prompts contained the fully specified query
including patient context. The food-safety comparator results therefore
represent context-available prompting controls, not matched-input ablations of
the clarification procedure.

Prompt templates and output schemas were fixed across models within each task
and configuration. Verbatim prompts, schemas, decoding settings, retry rules,
parsers, prompt hashes and task-specific access timestamps are retained in the
released configuration manifests; their reporting structure is summarized in
Supplementary Methods.

\subsubsection{Statistical analysis}
\label{sec:statistics}

\paragraph{Response accuracy.}

Metrics were calculated at the case--query level from frozen row-level
predictions, using the task-specific outcome definitions given above.
Configurations were aligned by benchmark instance. The primary comparisons
were descriptive; no null-hypothesis significance tests or multiplicity
adjustments were applied. Each model--configuration result was reported with
its analysis denominator, missing outputs and abstentions. Analyses restricted
to answered cases were designated as selective and were reported with
coverage. Direct comparisons used either the complete cohort or an explicitly
defined common-case subset.

\paragraph{Between-model consistency.}

To assess whether performance depended on the downstream language model, each
case-level metric was first calculated separately for each of the five
downstream models under each configuration. For every configuration and
metric, the five model-specific scores were then summarized by their mean,
population standard deviation and range. The mean represented average
performance across the model panel, and the standard deviation summarized how
widely the five model-specific scores were distributed around that mean; a
smaller value indicated that the five models produced more similar results
under that configuration. Between-model variation was calculated across the
five models, could not be calculated for one model alone and did not measure
change over time or within an individual model. Because a small standard
deviation is uninformative when all five models perform similarly poorly, it
was interpreted together with the corresponding mean. Diagnosis metrics were
computed over all 1,034 cases with abstentions counted as incorrect and, as a
robustness check, over the 999 cases answered by the \AMA, using the same
cases for all three configurations; dietary metrics were computed over all 487
queries. Exact Top-1 diagnosis accuracy was reported separately because its
near-zero baseline means made dispersion alone misleading. The five models
were purposively selected; these summaries were therefore descriptive and were
not interpreted as estimates for a population of language models.

\paragraph{Uncertainty under repeated generation.}
\label{sec:stability-analysis}

Repeated generation was examined as an exploratory analysis of output
stability. Two evaluation cases were selected from each task, and each
model--configuration pair was sampled independently 500 times per case. In the
\AMA condition, the recorded clarification trajectory, terminal state and
clarified query were held fixed so that only downstream response generation
was resampled; in the \textsc{Basic} and \textsc{RaR} conditions, the complete
response was regenerated on every repetition. The resulting values therefore
characterize downstream output dispersion under these two protocols rather
than a symmetric measure of end-to-end uncertainty.

For observed outputs with empirical frequencies \(p(y)\), predictive entropy
was
\begin{equation}
 H(Y)=-\sum_y p(y)\log_2 p(y).
 \label{eq:predictive-entropy}
\end{equation}
For diagnosis, three quantities were reported per case: \(K\), the number of
distinct normalized exact diagnosis strings returned across the 500
repetitions; \(H_{\mathrm{exact}}\), the entropy over those strings; and
\(H_{\mathrm{cat}}\), the entropy after mapping each string to its clinical
category, which separates variation in wording from variation in the clinical
conclusion. Outputs that could not be mapped to a category were retained as
distinct categories rather than excluded. For food safety, entropy was
calculated over the two normalized binary labels separately for the
\texttt{NOT\_OKAY}-labelled query (\(H_{\mathrm{unsafe}}\)) and the
\texttt{OKAY}-labelled query (\(H_{\mathrm{safe}}\)), and the mean entropy was
the arithmetic mean of the two. Because the two food-safety queries had
opposite reference labels, their 500 outputs each were additionally pooled
into a balanced set of 1,000 decisions per model--configuration pair, from
which accuracy, MCC, recall, precision and false-positive rate were
calculated. Lower entropy indicated greater output consistency but did not
indicate correctness; accuracy was assessed separately and did not enter the
entropy calculations. Because only two cases per task were evaluated, these
analyses were interpreted as case studies rather than population-level
stability estimates.

\subsubsection{Reproducibility and ethical considerations}
\label{sec:reproducibility}

No LLM was trained or fine-tuned, and no GPU was required to recalculate the
reported metrics from the frozen outputs. Commercial model calls cannot be
reproduced byte-for-byte because provider infrastructure and served aliases
may change. We therefore froze the requested model identifiers, prompts,
structured evidence, outputs and SHA-256 hashes used for the reported
analyses. The offline workflow verifies these artifacts and recalculates the
tables without network, API or database access. A new live model call is
treated as a replication attempt rather than exact regeneration.

\label{sec:ethics}

The study involved no participant recruitment, clinical intervention or
deployment in patient care. The diagnosis task used synthetic patient records,
and the oracle responses in both tasks were simulated from fields already
contained in the benchmarks. The evaluation therefore does not establish
clinical safety, effectiveness, fairness or usability with real patients. The
\AMA should not be interpreted as an autonomous diagnostic or dietary
decision system.

\section*{Data availability}

The symptom--diagnosis benchmark was derived from synthetic Synthea records and
contains no real patient records. The food-safety benchmark was constructed for
this study from authored dietary-query templates and the rule sources described
in the manuscript. Frozen benchmark inputs and row-level predictions will be
made available with the peer-review release, subject to confirmation of the
final redistribution and citation metadata. UMLS Metathesaurus source files are
available separately from the US National Library of Medicine under the UMLS
license and are not redistributed by the authors.

\section*{Code availability}

Source code for constructing the knowledge graphs, running the ambiguity
experiments and reproducing the reported metrics is available at https://github.com/HealthSciTech/ambiguity-mitigator-agent. Licensed UMLS files and commercial API credentials are not included.

\section*{Acknowledgements}
This research received no specific grant from any funding agency in the
public, commercial or not-for-profit sectors.

\bibliographystyle{naturemag}
\bibliography{ref}

% =====================================================================

\section*{Author contributions}
\addcontentsline{toc}{section}{Author contributions}
M.A. conceived the study, co-designed the architecture of the Ambiguity
Mitigation Agent, performed the analysis of the results and co-wrote the
manuscript. S.A.F. led the methodological and technical design of the Ambiguity Mitigation Agent and implemented it in openCHA; built the DxSxKG and FoodSafetyKG knowledge graphs; developed the symptom–diagnosis benchmark; conducted the experiments, computed the evaluation metrics and prepared the results tables; and drafted the original Methods and Supplementary Methods. M.A. and S.A.F. had equal contribution in this work. A.I.
curated the benchmark datasets, developed the Ambiguity Injection Module
and drafted the related-work material. I.A. contributed in the conceptualization and co-wrote the manuscript. I.A.,
H.C., R.J. and A.M.R. supervised the project and critically revised the
manuscript. All authors reviewed and approved the final version.
% TODO-AUTH  npj requires an explicit statement, e.g.:
% M.A. conceived the study, designed the framework and wrote the
% manuscript. ... All authors reviewed and approved the final manuscript.

\section*{Competing interests}
\addcontentsline{toc}{section}{Competing interests}
The authors declare no competing interests.

% =====================================================================

% #####################################################################
%                        SUPPLEMENTARY INFORMATION
% #####################################################################
\clearpage
\appendix
\setcounter{section}{0}
\setcounter{subsection}{0}
\setcounter{table}{0}
\setcounter{figure}{0}
\renewcommand{\thesection}{\Alph{section}}
\renewcommand{\thesubsection}{\thesection.\arabic{subsection}}
\renewcommand{\theHsection}{supp.\arabic{section}}
\renewcommand{\theHsubsection}{\theHsection.\arabic{subsection}}
\renewcommand{\thetable}{\arabic{table}}
\renewcommand{\thefigure}{\arabic{figure}}
\renewcommand{\figurename}{Supplementary Fig.}
\renewcommand{\tablename}{Supplementary Table}

\providecommand{\tablenotemark}[1]{\textsuperscript{#1}}
\providecommand{\tablenotetext}[2]{%
  \par\smallskip\noindent\textsuperscript{#1}\,#2}

\definecolor{SIPromptBack}{HTML}{F3F7FA}
\definecolor{SIPromptFrame}{HTML}{7A9CB3}
\definecolor{SIPromptTitle}{HTML}{244B63}
\newsavebox{\sippromptboxsave}
\newenvironment{sippromptbox}[1]{%
  \par\medskip\noindent
  \begin{lrbox}{\sippromptboxsave}%
  \begin{minipage}{0.94\linewidth}%
  \textbf{\color{SIPromptTitle}#1}\par\smallskip
}{%
  \end{minipage}%
  \end{lrbox}%
  \fcolorbox{SIPromptFrame}{SIPromptBack}{\usebox{\sippromptboxsave}}%
  \par\medskip
}

\section{Supplementary Methods}
\label{app:supplementary-methods}

These Supplementary Methods provide the implementation, provenance and prompt
details required to interpret and reproduce the experiments reported in the
main manuscript. The analysis cohorts, task definitions and primary outcomes
are identical to those in the main Methods. Licensed UMLS files and commercial
API credentials are not redistributed.

\subsection{Diagnosis benchmark construction}

The diagnosis benchmark was generated with Synthea
\cite{walonoski2018synthea}, using commit \texttt{c807432}, Florida as the geographic
setting, ages 0--90 years, all disease modules, a 10-year history and a
reference date of 24 August 2025. CSV and symptom exporters were enabled. The
patient-generation seed was 404, the clinician seed was 1755999126513 and the
final run was recorded at \texttt{2025-08-24T01:32:06Z}. It used Java 17.0.14
and generated 4,884 synthetic patients.

Benchmark construction retained disorder-labelled pathology episodes with at
least one symptom. Parenthetical Synthea qualifiers were removed from labels,
and duplicate records were collapsed to one case--diagnosis instance for each
patient and retained pathology. Forty pathologies were retained across the
complete benchmark. Of the generated patients, 4,518 contributed at least one
instance; 366 contributed none and were excluded.

Patients were partitioned before evaluation. The training, validation and
test partitions contained 3,163, 904 and 451 patients, respectively, with no
patient overlap. For each instance, a non-empty subset of symptoms was withheld
while at least one symptom remained visible. The test partition contained
1,034 instances and 32 reference diagnoses. Its 8,503 symptom occurrences
comprised 4,299 visible occurrences (50.6\%) and 4,204 withheld occurrences
(49.4\%).

\begin{table}[htbp]
\centering
\caption{Flow of synthetic patients and case--diagnosis instances into the
diagnosis benchmark.}
\label{tab:supp-synthea-flow}
\small
\begin{tabular}{lrr}
\toprule
Stage or partition & Patients & Instances \\
\midrule
Generated by Synthea & 4,884 & -- \\
Excluded: no retained benchmark instance & 366 & -- \\
Included in the frozen benchmark & 4,518 & 10,374 \\
\quad Training partition & 3,163 & 7,258 \\
\quad Validation partition & 904 & 2,082 \\
\quad Test partition & 451 & 1,034 \\
\bottomrule
\end{tabular}
\end{table}

The partition-file SHA-256 hashes were:
\begin{flushleft}\small
Training: \nolinkurl{51960aa4b1fbf6a94bd0bd72c6cbab65fd4c04b313768d0d4a259ca84d990573}\\
Validation: \nolinkurl{7eeb5a0226f39834a3ca74a321e73fbc0e5a01ab747c981d17920be9dc5c0fa3}\\
Test: \nolinkurl{c2eaa6dd06ca784e5748344ed6f4be43a2925eecdce166d2ef48db91028eeb28}.
\end{flushleft}
These files preserve the evaluated assignments and symptom partitions even if
the original randomization is not rerun.

\clearpage
\subsection{DxSxKG construction and concept alignment}

DxSxKG was constructed from UMLS 2025AA \cite{bodenreider2004unified}, using
\texttt{MRCONSO} for names and synonyms, \texttt{MRSTY} for semantic types and
\texttt{MRREL} for relations. The held-out test partition did not contribute
graph relations. Each node was keyed by its concept unique identifier (CUI)
and retained its preferred name, synonyms, semantic type and source vocabulary.

Diagnosis nodes were concepts assigned T047 (Disease or Syndrome) or T048
(Mental or Behavioral Dysfunction). Symptom nodes were concepts assigned T184
(Sign or Symptom) or T033 (Finding). Concepts satisfying both definitions were
represented as diagnoses. Direct diagnosis--manifestation relations were
normalized to \texttt{HAS\_SYMPTOM}; reverse-direction relations were
reoriented, suppressed records were excluded and duplicate pairs were
collapsed. Original relation labels were retained as provenance.

Clinical-family relations were stored as \texttt{IS\_A} edges. They supported
only the family-level terminal state and did not affect candidate retrieval,
question selection or belief updating.

\begin{table}[htbp]
\centering
\caption{DxSxKG artifact used in the reported evaluation.}
\label{tab:supp-dxsxkg}
\small
\begin{tabular}{lr}
\toprule
Component & Count \\
\midrule
Diagnosis nodes & 129,815 \\
Symptom nodes & 340,842 \\
\texttt{HAS\_SYMPTOM} relations & 82,813 \\
\texttt{IS\_A} relations & 1,400 \\
\bottomrule
\end{tabular}
\end{table}

Visible symptoms were normalized by lowercasing, collapsing whitespace and
removing parenthetical qualifiers. Resolution then attempted exact-name,
stored-synonym and full-text matching. If these steps failed, the inner model
could propose a UMLS label, which the controller accepted only when it exactly
matched an eligible graph node. Unresolved strings did not contribute to
candidate retrieval. The reference diagnosis and withheld symptoms were
unavailable during resolution.

The UMLS archive was \texttt{umls-2025AA-metathesaurus-full.zip}; its source
files were timestamped 28 April 2025. Their SHA-256 hashes were:
\begin{flushleft}\small
\texttt{MRCONSO.RRF}: \nolinkurl{bfb75116adc9f0308025ea90c05b5c356896198d307687b820979e6bf782f058}\\
\texttt{MRREL.RRF}: \nolinkurl{44d896e9ff8046b38a0ebd15e40615314e12ef37d0d99901bb3976014b1cbfd7}\\
\texttt{MRSTY.RRF}: \nolinkurl{2712ae33d6450492a0415ea9d3291dc6cee6e0b19b89df7081044515b69f227a}.
\end{flushleft}

\clearpage
\subsection{Diagnosis controller implementation}

The frozen controller was identified as
\nolinkurl{diagnostic-agent-release-v1-compat-2026-07}. A short or
ambiguity-flagged mention could retain up to three graph resolutions; another
mention retained the highest-ranked resolution. The inner model then made one
expansion decision and could propose up to three additional symptom terms. A
proposed term entered retrieval only after resolution to an eligible graph
concept. Expansion was not repeated and could not access withheld symptoms.

Let \(\mathcal P\) denote the retained concepts. A diagnosis required one
match when \(|\mathcal P|=1\) and at least two otherwise. The graph returned at
most eight diagnoses, ordered by decreasing match count, decreasing match
fraction and increasing number of connected symptoms. Fixed diagnosis-name
exclusions were then applied without replacement.

The controller assigned equal initial weight to each diagnosis. At each turn
it selected the unasked symptom with greatest positive expected information
gain among diagnoses with non-zero posterior weight. The oracle answered from
the complete synthetic symptom profile. A positive answer retained diagnoses
linked to the symptom; a negative answer removed them. Remaining weights were
renormalized.

Clarification stopped when one diagnosis exceeded posterior probability 0.95,
no question had positive information gain or five questions had been asked. A
unique remaining diagnosis produced \textsc{Answer}. Otherwise, an eligible
immediate \texttt{IS\_A} family covering at least
\(\lceil|\mathcal H_0|/2\rceil\) initially retained diagnoses produced
\textsc{Partial}; absence of such a family produced \textsc{Abstain}.

\clearpage
\subsection{Food-safety benchmark construction}

The food-safety benchmark contained 487 instances: 301 labelled
\texttt{OKAY} and 186 labelled \texttt{NOT\_OKAY}. Each record preserved the
fully specified query, ambiguous query, removed patient context, reference
label and reference explanation. In the Agent condition, the removed context
was accessible only to the oracle and was not included in food extraction,
graph expansion, question selection or downstream input before disclosure
through an answer.

\clearpage
\subsection{FoodSafetyKG construction and provenance}

FoodSafetyKG integrated the USDA FoodData Central full CSV release dated 18
December 2025 \cite{usda_fdc}, UMLS 2025AA condition concepts
and hierarchies, and clinical-guideline rules. FoodData Central supplied food
composition and nutrient records. UMLS supplied normalized condition concepts,
aliases and parent--child relations. The guideline registry supplied food
properties, thresholds and condition-specific risk or safety associations.
Each authored or derived relation retained its source and evidence type.

The graph represented food phrases, USDA food items, ingredients, nutrients,
food properties, conditions, condition families and aliases. Its principal
evidence paths connected food phrases through ingredients and properties to
conditions, and food items through nutrients or properties to conditions.
Materialized \texttt{RISKY\_FOR} and \texttt{SAFE\_FOR} relations shortened
these paths without discarding provenance. The complete FoodSafetyKG was
available throughout the reported evaluation.

\begin{table}[htbp]
\centering
\caption{FoodSafetyKG components used in the reported evaluation.}
\label{tab:supp-foodkg}
\small
\begin{tabular}{lr}
\toprule
Component & Count \\
\midrule
USDA food-item nodes & 2,085,340 \\
Nutrient nodes & 477 \\
Food-phrase nodes & 298 \\
Ingredient nodes & 130 \\
Food-property nodes & 60 \\
Condition nodes & 67 \\
Condition-family nodes & 16 \\
Condition-alias nodes & 138 \\
\texttt{HAS\_NUTRIENT} relations & 27,094,028 \\
Derived food-item \texttt{HAS\_PROPERTY} relations & 2,542,284 \\
Materialized food-item risk relations & 21,030,444 \\
Curated food-phrase--ingredient relations & 398 \\
Curated ingredient--property relations & 341 \\
Property--condition \texttt{RISKY\_FOR} relations & 126 \\
Property--condition \texttt{SAFE\_FOR} relations & 67 \\
\bottomrule
\end{tabular}
\end{table}

The FoodData Central files were extracted from
\texttt{FoodData\_Central\_csv\_2025-12-18.zip}. Their SHA-256 hashes were:
\begin{flushleft}\small
\texttt{food.csv}: \nolinkurl{a1ec0c8ebee04f91de8c9549486f555f3a8367e538a7a6da4dbb0f068f76a983}\\
\texttt{food\_nutrient.csv}: \nolinkurl{e4085bacf4e986304f01545e90a2b6ab5a30f204395a471db4d9854cd8cd74ed}\\
\texttt{nutrient.csv}: \nolinkurl{226ba937d1a73e8c87bd7fdcdd577524699bef4025c9232ee474b7d3083498ff}.
\end{flushleft}

The rule registry grouped its principal clinical sources into ADA, AHA, ACG,
NKF KDOQI, ACR, FARE, ATA, ACOG, NOF, AASLD, AHS, FDA, WHO and ESC guideline
families. The machine-readable registry records issuing body, year, source URL,
evidence level and affected property--condition relation.

\clearpage
\subsection{Food-safety controller implementation}

The frozen controller was identified as
\nolinkurl{food-agentic-pomdp-v4-shared-stopping}. The inner model extracted
one or more food terms from the visible query; a deterministic extractor
provided a fallback term. The controller queried direct
\texttt{RISKY\_FOR} relations first and used the permitted
food--ingredient--property--condition path when direct retrieval returned no
condition. Results from all accepted terms were combined, deduplicated and
limited to 50 conditions.

The inner model then made one expansion decision and could propose up to three
additional food, ingredient, synonym or category terms. A proposed term
affected retrieval only after graph resolution and return of at least one
supported risk condition. Expansion was performed once and could not access
the hidden patient context. It added no condition in the 487 frozen cases.

Each condition received equal initial weight. Condition families reachable
through one to three \texttt{IS\_A} edges defined family questions; individual
conditions defined leaf questions. The controller selected the question with
greatest expected information gain. Exact ties favoured a family question and
then the normalized question name. When one unasked condition remained, the
controller asked it directly.

Positive family answers retained member conditions; negative answers removed
them. Leaf answers retained or removed one condition. Clarification ended
after confirmation of a risk condition, denial of all graph-supported risks,
posterior probability greater than 0.95 or exhaustion of positive-information-
gain questions. There was no numerical turn limit; the finite question pool
bounded the interaction.

After clarification, the controller retrieved supporting graph paths and
checked applicable \texttt{SAFE\_FOR} relations. A confirmed risk without a
safety exception produced a symbolic \texttt{NOT\_OKAY} hint. Denial of all
retrieved risks or an applicable safety exception produced an \texttt{OKAY}
hint. Incomplete clarification produced no symbolic class. The inner model
could organize the evidence record but could not change its symbolic fields.

The frozen Stage-A artifact contained all 487 cases and had SHA-256 hash
\begin{center}\small
\nolinkurl{ef34b7872048178e313d52d730228322e4328b9117b7fe1a128b95b25899d1cc}.
\end{center}
Clarification completed in 485 cases; two had no graph-supported initial
hypothesis. The mean was 5.96 questions per case and the maximum was 11. Every
case was passed to the downstream model for one binary prediction.

\clearpage
\subsection{Model routing and inference configuration}

The inner clarification model was GPT-5.5 through the OpenAI API. The same five
downstream models were used in both tasks. Model developer and API provider
were recorded separately because Gemini and Llama were served through
OpenRouter.

\begin{table}[htbp]
\centering
\caption{Model routing used for the frozen evaluation.}
\label{tab:supp-models}
\scriptsize
\begin{tabularx}{\textwidth}{@{}p{2.3cm}p{1.7cm}p{1.8cm}>{\raggedright\arraybackslash}X@{}}
\toprule
Display label & Developer & API provider & Requested identifier \\
\midrule
GPT-5.5 & OpenAI & OpenAI & \texttt{gpt-5.5} \\
Claude Opus 4.8 & Anthropic & OpenRouter & \texttt{claude-opus-4-8} \\
Gemini 3.1 Pro & Google & OpenRouter & \texttt{google/gemini-3.1-pro-preview} \\
Llama 3.3 70B Instruct & Meta & OpenRouter & \texttt{meta-llama/llama-3.3-70b-instruct} \\
Mistral Large & Mistral AI & OpenRouter & \texttt{mistral-large-2512} \\
\bottomrule
\end{tabularx}
\end{table}

Prompts and output schemas were fixed within each task and configuration.
Provider-returned identifiers, request timestamps, output-token limits, retry
outcomes, raw outputs and parsed outputs were stored when available. The
frozen food manifests record runs for Claude, Gemini, Llama and Mistral on 20
July 2026 (UTC). A new commercial-model call is a replication rather than an
exact regeneration because a served alias or stochastic output can change.

For diagnosis, all configurations received the same incomplete symptom
presentation. For food safety, the frozen Direct and RaR prompts contained the
fully specified query, including patient context, whereas the Agent began with
the ambiguous query. The food baselines are therefore context-available
prompting controls.

\clearpage
\section{Prompt templates and output schemas}
\label{supp:prompts}

The templates below document the instruction roles, decision constraints and
output schemas that affected the evaluated outputs, allowing reviewers to
separate model behaviour from deterministic graph operations.
Angle-bracketed fields were populated at runtime. Exact machine-readable strings,
model identifiers, decoding settings and prompt hashes should accompany the
released evaluation records. Deterministic graph queries, information-gain
calculations and belief updates did not use prompts.

\subsection{Direct and Rephrase-and-Respond prompts}

\begin{sippromptbox}{Supplementary Prompt 1 | Direct diagnosis output}
\small\ttfamily\raggedright
\textless INCOMPLETE SYMPTOM CONVERSATION\textgreater\\[2pt]
You are a medical assistant helping to determine a likely diagnosis.\\
Provide your top 5 possible diagnoses ranked from most likely to least likely.\\
Return ONLY valid JSON with this exact schema:\\
\{``diagnoses'': [\{``name'': ``diagnosis name'', ``icd10'': ``ICD-10-CM code''\}, ...]\}\\
Provide exactly 5 entries. For each, give the diagnosis name and its standard
ICD-10-CM code.
\end{sippromptbox}

\begin{sippromptbox}{Supplementary Prompt 2 | Direct food-safety output}
\small\ttfamily\raggedright
\textless FULLY SPECIFIED FOOD QUERY\textgreater\\
The question asks whether it is OK for the patient to eat/drink/do the thing
described, given their context. Answer with exactly one token: OK or NOT\_OK.
Do not explain.
\end{sippromptbox}

RaR used two calls. The first received:
\begin{sippromptbox}{Supplementary Prompt 3 | RaR reformulation call}
\small\ttfamily\raggedright
``\textless SUPPLIED QUERY\textgreater''\\
Rephrase and expand the question, and respond.
\end{sippromptbox}
The second call received:
\begin{sippromptbox}{Supplementary Prompt 4 | RaR answer call}
\small\ttfamily\raggedright
(original) \textless SUPPLIED QUERY\textgreater\\
(rephrased) \textless REFORMULATED QUERY\textgreater\\
Use your answer for the rephrased question to answer the original question.\\
\textless TASK-SPECIFIC OUTPUT INSTRUCTION FROM PROMPT 1 OR 2\textgreater
\end{sippromptbox}

\subsection{Diagnosis inner-model prompts}

\begin{sippromptbox}{Supplementary Prompt 5 | UMLS fallback resolver}
\small\ttfamily\raggedright
You are a medical terminology expert.\\
Map this patient symptom to the closest UMLS concept name in a medical
knowledge graph.\\
Patient symptom: ``\textless SYMPTOM\textgreater''\\
Sample concept names from the knowledge graph:\\
\textless UP TO 40 GRAPH LABELS\textgreater\\
Rules:\\
1. Return ONLY the concept name -- no explanation.\\
2. If the symptom matches a sample name exactly, return that.\\
3. Otherwise return the closest medical equivalent.\\
4. Keep it under 6 words.
\end{sippromptbox}

\begin{sippromptbox}{Supplementary Prompt 6 | Diagnosis question wording}
\small\ttfamily\raggedright
Rephrase the medical symptom ``\textless GRAPH SYMPTOM NAME\textgreater'' as a
simple yes/no question for a patient. Return only the question. Under 15 words.
\end{sippromptbox}

\subsection{Diagnosis downstream-model prompts}

\begin{sippromptbox}{Supplementary Prompt 7 | Diagnosis \textsc{Answer} state}
\small\ttfamily\raggedright
You are a board-certified physician acting as a diagnostic verifier.\\
The symbolic POMDP agent selected one diagnosis with high confidence.\\
Verify it is clinically plausible given the evidence summary.\\
If plausible: confirm it.\\
If NOT plausible: choose the best alternative from the candidate set ONLY.\\
If the top candidate is in the candidate set and matches the evidence, always
confirm it.\\
Return ONLY valid JSON:\\
\{``final\_diagnosis'': ``exact candidate name'', ``decision'': ``confirmed'',
``prediction\_top5'': [``d1'',``d2'',``d3'',``d4'',``d5''], ``reason'': ``one
evidence-based sentence''\}
\end{sippromptbox}

\begin{sippromptbox}{Supplementary Prompt 8 | Diagnosis \textsc{Partial} state}
\small\ttfamily\raggedright
You are a board-certified physician acting as a diagnostic ranker.\\
The symbolic agent identified a clinical family but cannot distinguish between
candidates. Rank the candidates using the confirmed and denied evidence.\\
Rules:\\
- Choose ONLY from the candidate set.\\
- Use confirmed and denied symptoms as primary evidence.\\
- Use prevalence only as a tiebreaker.\\
- Prefer viral to bacterial sinusitis unless purulent discharge, duration over
7 days, severe focal facial pain or double worsening is present.\\
- When diarrhea, productive cough or wheezing is confirmed, prefer supported
lower-respiratory diagnoses over upper-respiratory diagnoses.\\
- When symptoms are limited to throat, tonsils or mild nasal symptoms, prefer
supported upper-respiratory diagnoses.\\
Return ONLY valid JSON with \texttt{final\_diagnosis}, a five-item
\texttt{prediction\_top5} and one evidence-based \texttt{reason}.
\end{sippromptbox}

The post-abstention prompt supported counterfactual inspection only. Its output
was not scored as an answered framework case.

\clearpage
\subsection{Food-safety inner-model prompts}

\begin{sippromptbox}{Supplementary Prompt 9 | Food-query ambiguity analysis}
\small\ttfamily\raggedright
You are a clinical food-query ambiguity analyst. Analyse only the visible
patient query. Do not invent a diagnosis or restriction.\\
Patient query: ``\textless VISIBLE QUERY\textgreater''\\
Check for food-identity, ingredient, preparation, portion, patient-context and
timing ambiguity.\\
Provide simple lowercase food, ingredient and useful generic-category terms
for KG retrieval. Include compound and component terms when both are visible,
but do not add foods not mentioned.\\
Return ONLY this JSON:\\
\{``ambiguity\_types'': [``type''], ``food\_terms'': [``term''],
``missing\_context'': [``item''], ``multi\_interpretation\_needed'':
[``term''], ``summary'': ``one sentence''\}
\end{sippromptbox}

\begin{sippromptbox}{Supplementary Prompt 10 | One-time food-term expansion}
\small\ttfamily\raggedright
You are evaluating a food-safety KG hypothesis set.\\
Visible query: \textless VISIBLE QUERY\textgreater\\
Ambiguity report: \textless AMBIGUITY JSON\textgreater\\
Food terms already searched: \textless TERMS\textgreater\\
Current risky-condition hypotheses: \textless CONDITIONS\textgreater\\
Expand if the set is empty, has fewer than three hypotheses, or is clinically
incoherent with the visible food. Suggest at most three food, ingredient,
synonym or generic-category search terms. Do not suggest diagnoses and do not
use undisclosed patient context.\\
Return ONLY this JSON:\\
\{``needs\_expansion'': false, ``reason'': ``one sentence'',
``additional\_terms'': []\}
\end{sippromptbox}

\begin{sippromptbox}{Supplementary Prompt 11 | Food question wording}
\small\ttfamily\raggedright
For a leaf condition: Rephrase the medical condition name into a simple yes/no
patient question. Return ONLY the question, without explanation.\\
For a family: Rephrase this condition-family screen into a simple yes/no
patient question asking whether the patient has any condition in that family.
Return ONLY the question, without explanation.
\end{sippromptbox}

\begin{sippromptbox}{Supplementary Prompt 12 | Structured food evidence record}
\small\ttfamily\raggedright
You are a clinical evidence synthesiser. Build the structured food-safety
evidence bundle u-prime for a final model. Use only the visible query,
disclosed clarification answers and supplied KG evidence. Do not infer an
undisclosed patient condition. Do not change clarification completeness or any
supplied symbolic binary hint.\\
Inputs: visible query; ambiguity report; food terms; disclosed Q\&A; remaining
conditions; risky and safe KG evidence; food properties; resolution status;
condition family; symbolic binary hint.\\
Return ONLY JSON with these keys: presentation, ambiguity\_resolved,
food\_interpretation, confirmed\_evidence, ruled\_out, never\_asked,
remaining\_uncertainty, resolution\_complete, condition\_family,
candidate\_conditions, food\_properties, symbolic\_binary\_hint and
safety\_question.
\end{sippromptbox}

Before downstream generation, the controller restored the symbolic values for
resolution completeness, condition family, candidate conditions, food
properties and the binary hint. The evidence-synthesis model could not alter
these fields.

\clearpage
\subsection{Food-safety downstream-model prompts}

\begin{sippromptbox}{Supplementary Prompt 13 | Completed food clarification}
\small\ttfamily\raggedright
You are a clinical dietitian verifying a binary food-safety decision. The
symbolic ambiguity agent asked clarification questions and completed its path.
Use only disclosed answers, the food query and supplied KG evidence. Do not
assume undisclosed conditions.\\
Distinguish direct contraindications from conditional or moderation risks.
Treat raw fish or undercooked meat in pregnancy, alcohol in pregnancy or liver
disease, gluten in celiac disease, lactose in lactose intolerance, strongly
high-sodium processed foods in kidney disease or hypertension, directly
applicable stimulant/caffeine or GERD triggers, and high-sugar drinks or
desserts in diabetes as direct risks when supported by the supplied evidence.\\
Do not treat a general property as an absolute contraindication when the food
wording, preparation or portion makes the relation conditional. Foods such as
rice, oatmeal, fruit, whole-grain bread and salmon may be acceptable in
moderation unless a direct relation applies to the exact item. Respect
modifiers such as gluten-free and low-sodium.\\
Return ONLY valid JSON:\\
\{``decision'': ``okay or not\_okay'', ``confidence'': ``high or medium or
low'', ``reasoning'': ``one evidence-based sentence''\}
\end{sippromptbox}

\begin{sippromptbox}{Supplementary Prompt 14 | Limited food clarification}
\small\ttfamily\raggedright
You are a clinical dietitian making the required binary food-safety judgment
after a bounded clarification process. Use only the visible query, disclosed
clarification answers, remaining KG hypotheses, ingredients, preparation and
food properties. Do not assume an undisclosed condition. Use low confidence
when essential information remains missing.\\
Return ONLY valid JSON:\\
\{``decision'': ``okay or not\_okay'', ``confidence'': ``high or medium or
low'', ``reasoning'': ``one sentence''\}
\end{sippromptbox}

\clearpage
\section{Outcome processing and reproducibility}
\label{supp:outcomes-reproducibility}

\subsection{Diagnosis outcomes}

Exact Top-1 and exact Recall@5 required literal agreement after case folding,
trimming and whitespace normalization. Secondary analyses resolved returned
and reference names to UMLS concepts and used their ICD-10 mappings. Broader
categories were derived from the UMLS hierarchy. The authors checked the final
mapping table before calculating the outcomes.

Selective outcomes used the 999 answered instances as the denominator and
were accompanied by coverage. Overall outcomes used all 1,034 test instances
and scored abstentions as incorrect. A diagnosis generated for counterfactual
inspection after \textsc{Abstain} was not counted as an answered output.

\subsection{Food-safety outcomes}

Raw \texttt{OK} and \texttt{NOT\_OK} tokens were normalized to
\texttt{OKAY} and \texttt{NOT\_OKAY}. \texttt{NOT\_OKAY} was the positive
class. All valid binary predictions entered the primary analysis, including
the two cases in which clarification ended with limited evidence. Missing or
invalid outputs were retained as missing and were not assigned a class.

\subsection{Frozen artifacts and offline verification}

The release contains the 30 primary prediction files formed by two tasks, five
downstream models and three configurations. A SHA-256 manifest records the
byte size and hash of each artifact. The offline workflow verifies the
manifest and recalculates the reported metrics without API, database, UMLS or
USDA access. Live runners write to a separate directory and do not overwrite
the frozen release.

Exact reproduction means recalculation from the frozen inputs and outputs. A
new commercial-model call is a replication because the provider can update a
served alias and stochastic generation can return another valid response.

\clearpage
\section{Repeated-generation analyses}
\label{supp:stability}

Two cases were selected from each task. Each model--configuration pair was
sampled 500 times per case. In the Agent condition, the recorded clarification
trajectory, terminal state and clarified query were held fixed; only the
downstream response was regenerated. Direct and RaR regenerated their complete
responses on each repetition.

For diagnosis, predictive entropy was calculated over normalized exact
diagnosis strings and, separately, mapped diagnostic categories. For food
safety, entropy was calculated over the two binary labels. For empirical
output frequencies \(p(y)\),
\begin{equation}
H(Y)=-\sum_y p(y)\log_2p(y).
\end{equation}
Lower entropy indicates greater consistency, not greater accuracy. These
analyses characterize four selected cases and do not estimate stability over
either complete benchmark. Because the Agent trajectory was fixed, they do not
measure end-to-end variability in question selection or simulated answers.

\clearpage
\section{Scope of the controlled evaluation}
\label{supp:scope}

The evaluation used synthetic or authored cases and simulated answers. It did
not test patient comprehension, contradictory or uncertain answers,
clinician--AI interaction, treatment recommendations, workflow integration or
prospective outcomes. The diagnosis task measured retrieval of a recorded
synthetic diagnosis from incomplete symptoms; the food task measured binary
classification within the represented food and condition relations. Neither
task evaluated an autonomous clinical decision system.

The framework inherits the coverage and accuracy limits of its graphs. A
missing entity or relation can prevent retrieval of a relevant hypothesis or
question. A supported relation can also be too general for an individual
because preparation, portion, disease severity, medication use and comorbidity
may alter its applicability. These limits motivate evaluation with
independently reviewed cases, noisy answers and naturally occurring queries.
 
%---------------------------------------------------------------------
\clearpage

%======================================================================
%  Bridge: source corpora -> AIM.
%  Short lead-in to be placed immediately BEFORE
%  \section{Ambiguity Injection Module} (\label{app:aim}) in main.tex.
%  Deliberately does NOT restate the detailed provenance/counts already
%  given in the Supplementary Methods benchmark-construction subsections
%  (Diagnosis benchmark construction; Food-safety benchmark construction;
%  DxSxKG / FoodSafetyKG construction). It only names the source corpora
%  with citations and frames AIM as the procedure that turned them into
%  the underspecified benchmarks.
%  Uses only keys defined in ref.bib. Defines NO label (app:aim lives in
%  main.tex); do NOT \input AIM_rewritten.tex alongside it (label clash).
%======================================================================

Both evaluation sets in this study were derived from established,
publicly documented sources rather than authored from scratch. The
symptom--diagnosis benchmark is built from Synthea synthetic patient
records~\cite{walonoski2018synthea,chen2019synthea}, with hypotheses and
clarification questions supported by a symptom--diagnosis graph normalised
against the UMLS Metathesaurus~\cite{bodenreider2004unified,nlmUMLS2025AA}.
The dietary-safety benchmark pairs food queries with patient-context health
factors, adjudicated through a curated food-safety graph that integrates
USDA FoodData Central~\cite{usda_fdc,fukagawa2022fdc}, UMLS condition
concepts~\cite{bodenreider2004unified,nlmUMLS2025AA}, and a registry of
clinical-guideline-derived rules. The exact generation settings, cohort
flow, knowledge-graph construction, and provenance for both benchmarks are
given in the benchmark-construction subsections of these Supplementary
Methods (\S\ref{app:supplementary-methods}).

In every case the source instance is \emph{fully specified}: the Synthea
record lists the complete symptom set behind its diagnosis, and each
dietary seed states the health context that determines its safety label. A
benchmark that tests whether an agent can recover \emph{missing} patient
context therefore requires a controlled, reversible way to make these seeds
underspecified while retaining the withheld information as a hidden
reference. That transformation---rather than any change to the source
labels---is what converts the corpora above into ambiguity-mitigation
benchmarks, and it is performed by the Ambiguity Injection Module, which we
describe next (\S\ref{app:aim}).

\section{Ambiguity Injection Module}
\label{app:aim}
%---------------------------------------------------------------------
 
% CHECK: This appendix reproduces your full AIM section (all three
% transformation tracks, both example tables, worked examples), cleaned
% for formatting and citations. Two framing edits were necessary to
% keep it consistent with the main text: (1) the title no longer says
% "training-time", and an opening sentence states which track was used
% for the reported benchmarks, because Methods 4.4 says no model was
% trained; (2) "your food-safety benchmark" -> "the food-safety
% benchmark". The mental-health examples are retained as requested but
% that domain is not evaluated in the paper -- expect a reviewer to
% ask; consider a one-line note that AIM is domain-general and the
% mental-health rows illustrate transferability only.
 
Clarification agents require many examples in which the user's surface
utterance is intentionally incomplete, yet the information needed to
resolve that incompleteness is known. We address this gap with the
\emph{Ambiguity Injection Module} (AIM), an offline generator that
transforms \emph{fully specified} seed queries into
\emph{underspecified} variants while retaining the original intent as
latent structure. Conceptually, AIM instantiates an \emph{inverse
slot-filling} process: standard natural-language understanding maps
text to filled slots; AIM begins from text that already encodes those
slots and applies controlled transformations that remove, blur, or
rephrase slot content on the surface while preserving a recoverable
target for clarification. The two evaluation sets in this study were
produced with the rule-based masking track described below, applied to
structured slots (Synthea symptom lists~\cite{walonoski2018synthea}
and the patient-context phrase of curated dietary queries); the
remaining tracks are part of the module and are described for
completeness.
% CHECK: confirm that only rule-based masking was applied to the
% evaluated sets (this is what Results 2.1 states). If generalization
% or persona rewriting was also applied, remove this sentence and
% update Results 2.1.
 
\subsection{Entity grounding}
AIM first identifies domain-relevant spans that can be manipulated as
slots. For mental-health--oriented text, we combine clinical NLP
tagging (scispaCy~\cite{Neumann2019}, medspaCy~\cite{Eyre2021}) with
lightweight pattern constraints so that symptoms, temporal cues,
severity markers, and related phrases can be addressed consistently.
For nutrition-oriented text, we parse ingredients, amounts, and units
from free-form recipe or meal descriptions. Where the seed is already
structured, as in the benchmarks used here, slot boundaries are taken
directly from the source fields. This layer does not perform
downstream diagnosis or safety reasoning; it only supplies reliable
anchors for ambiguity operations.
 
\subsection{Complementary ambiguity mechanisms}
AIM composes three families of transformations that target different
linguistic sources of underspecification.
 
\paragraph{Rule-based masking} simulates omission and abstraction:
entities or phrases are dropped, quantities are replaced with vague
scoping, and domain-specific combinations of cues can be removed
jointly. In mental-health scenarios, we distinguish
\emph{safety-preserving} transformations from naive deletion:
high-risk content is not simply erased when doing so would remove
clinically salient signals; instead, selected terms are mapped to
vaguer surface forms so that downstream models still encounter
risk-relevant language while remaining underspecified for safe
decision-making.
 
\paragraph{Ontological generalization} replaces specific expressions
with broader categories by traversing lexical hypernym structure
(WordNet~\cite{Miller1995}) over short hop distances, supplemented by
small domain lexicons where WordNet coverage is thin. This yields
imprecision that remains plausible in user language (e.g., a specific
food $\rightarrow$ a coarse food class) without collapsing to trivial
supersenses.
 
\paragraph{Persona-conditioned rewriting} uses a large language model
instructed to imitate hesitant or novice users; the model paraphrases
the already masked or generalized text into natural vague phrasing.
Generated candidates are filtered with conservative post-hoc
constraints (e.g., suppressing reintroduction of disallowed explicit
markers when such a policy is enabled) to reduce degenerate or overtly
clinical surface forms.
 
\subsection{Corpus expansion and robustness}
The full pipeline can be run in a \emph{multi-strategy} mode that
enumerates strategy combinations, yielding many ambiguous counterparts
per seed and diversifying the pattern of missing information. An
optional lightweight perturbation stage injects small character- and
word-level noise into surface strings so that the paired records are
not tied to perfectly clean synthetic typography.
 
\subsection{Interface to the clarification agent}
Each AIM instance emits a tuple linking an ambiguous user-facing query
to metadata naming what was removed or generalized and to the original
fully specified query (and, when available, task labels such as
diagnosis or safety class). In the present study, the withheld
component is exposed only to the simulation oracle that answers
clarification questions and to the evaluation scripts; it is never
shown to the agent or to the downstream language model. The same
tuples can also serve as paired supervision for learned clarification
policies, without changing the runtime definition of ambiguity used by
the deployed agent.
 
\subsection{Illustrative AIM transformations}
\label{app:aim-examples}
Tables~\ref{tab:aim-examples-mh} and~\ref{tab:aim-examples-food} show
representative \emph{seed} utterances (fully specified text) and
\emph{injected} surface forms produced by individual mechanisms in our
implementation. The \emph{Latent} column names the information that
remains available as hidden structure (e.g., original query,
structured labels, or explicit clarification targets), while the
user-facing string is deliberately underspecified. Persona-conditioned
rewrites and typo-level perturbations are stochastic; the
corresponding rows are marked \emph{illustrative} and show paraphrases
consistent with the intended user personas in our generator prompts.
 
\begin{table}[htbp]
\centering
\small
\setlength{\tabcolsep}{4pt}
\caption{Mental-health domain: example transformations produced by
AIM masking, lexical generalization, and one illustrative vague user
paraphrase. \emph{Latent} indicates what is retained as hidden
structure and is not shown to the user at inference time.}
\label{tab:aim-examples-mh}
\begin{tabular}{p{0.19\linewidth} p{0.26\linewidth} p{0.37\linewidth} p{0.11\linewidth}}
\toprule
Mechanism & Seed (specific) & Surface after AIM & Latent \\
\midrule
Severity masking & I'm feeling \emph{extremely} anxious and having panic attacks for 2 weeks. I don't know what to do. & I'm feeling anxious and having panic attacks for 2 weeks. I don't know what to do. & full seed \\
\addlinespace
Temporal masking & I've been depressed \emph{for 2 weeks}. & I've been depressed for a while. & full seed \\
\addlinespace
Euphemism mapping (risk-preserving vagueness) & I've been having \emph{thoughts of hurting myself}. & I've been having thoughts.$^{a}$ & full seed \\
\addlinespace
Diagnostic phrase removal & I was diagnosed with \emph{major depressive disorder}. & I was diagnosed with.$^{b}$ & full seed \\
\addlinespace
Ontological generalization (domain map + WordNet) & After masking: \ldots\ and having \emph{panic attacks}\ldots & \ldots\ and having \emph{mental condition}\ldots\ (domain hypernym map where applicable) & masked intermediate \\
\addlinespace
LLM persona rewrite (illustrative) & (same seed as row~1, after rule-based steps) & ``I've been kind of on edge and my body does that thing where it freaks out, it's been going on for a bit.'' & full seed + strategy tags \\
\bottomrule
\end{tabular}
\par\smallskip
\raggedright\footnotesize
$^{a}$High-risk phrasing is mapped to a vaguer surface form rather than
silently deleted, so that the risk cue survives.
$^{b}$Diagnostic removal can leave minor grammatical artifacts; in
corpus construction such fragments are optionally post-filtered or
repaired. Each surface string is paired with the intact seed.
\end{table}
 
\begin{table}[htbp]
\centering
\small
\setlength{\tabcolsep}{4pt}
\caption{Food/dietary domain: examples of omission, abstraction, and
generalization that force clarification of health context (e.g.,
diabetes, allergy), amounts, or ingredients. The \emph{Latent} column
names the hidden structure that must be recovered before a safety
decision.}
\label{tab:aim-examples-food}
\begin{tabular}{p{0.19\linewidth} p{0.26\linewidth} p{0.37\linewidth} p{0.11\linewidth}}
\toprule
Mechanism & Seed (specific) & Surface after AIM & Latent \\
\midrule
Disease/condition masking & Can I eat white rice with chicken curry? \emph{I have type 2 diabetes.} & Can I eat white rice with chicken curry? & condition + safety label \\
\addlinespace
Allergen masking & I need a recipe with 2 cups flour and chicken, but \emph{I have a peanut allergy.} & I need a recipe with 2 cups flour and chicken, but. & allergen + full seed \\
\addlinespace
Quantity removal & Add \emph{2 cups} of flour to the bowl. & Add some of flour to the bowl. & quantities + full seed \\
\addlinespace
Ingredient omission & I need chicken, onions, and garlic for the stir-fry. & I need onions, and garlic for the stir-fry. (one ingredient dropped) & omitted slots + full seed \\
\addlinespace
Generalization only & I want grilled \emph{salmon} for dinner. & I want grilled \emph{fish} for dinner. & full seed \\
\addlinespace
LLM persona rewrite (illustrative) & (masked recipe-style request) & ``I want to make that creamy spicy thing, maybe with some meat? I can't remember what I bought.'' & full seed + variables \\
\addlinespace
Surface perturbation (illustrative) & Can I eat oatmeal with honey in the morning? & Can I eat oatmea with honney in the morning? & same as clean twin \\
\bottomrule
\end{tabular}
\end{table}

\section*{Declaration: use of large language models}
\addcontentsline{toc}{section}{Declaration: use of LLMs}
In preparing this manuscript, we used large language models (LLMs) to assist with drafting and editing portions of the text for clarity, as well as to support code development. All LLM-generated content was subsequently reviewed, verified, and revised by the authors. No LLM was used to generate scientific conclusions, and all references were manually verified for accuracy. The authors take full and sole responsibility for the accuracy and integrity of all content presented in this paper, in accordance with the journal policy on the use of large language models.
% TODO-LLM  Reworded from "conference 2026 LLM use policy" to journal policy.

\end{document}